\PassOptionsToPackage{table,xcdraw}{xcolor}
\documentclass[acmlarge,nonacm]{acmart}
\usepackage{multirow}
\usepackage{pifont}
\usepackage{colortbl}
\usepackage{amsmath}
\usepackage{makecell}
\usepackage{graphicx}
\usepackage{subcaption}

\AtBeginDocument{%
  }

\setcopyright{acmcopyright}
\copyrightyear{2026}
\acmYear{2026}
\acmDOI{}

\acmJournal{TOMM}

\begin{document}

%%
%% The "title" command has an optional parameter,
%% allowing the author to define a "short title" to be used in page headers.
\title{SWT: Self-Supervised Video Object Segmentation via \textbf{S}liding, \textbf{W}avelet and \textbf{T}ransportation}

%%
%% The "author" command and its associated commands are used to define
%% the authors and their affiliations.
%% Of note is the shared affiliation of the first two authors, and the
%% "authornote" and "authornotemark" commands
%% used to denote shared contribution to the research.
\author{Zhengtong Zhu}
\affiliation{%
  \institution{School of Computer Science and Technology, Soochow University}
  \country{China}
}
%\email{email: jjfang@stu.suda.edu.cn}

\author{Jiaqing Fan}
\authornote{Corresponding author}
\affiliation{%
  \institution{School of Computer Science and Technology, Soochow University}
  \country{China}
}
\email{email: jqfan@suda.edu.cn}
%\email{email: 20224027011@stu.suda.edu.cn}

%\author{Jiaqing Fan$^*$}
\author{Hanwen Qian}
\affiliation{%
  \institution{School of Computer Science and Technology, Soochow University}
  \country{China}
}

\author{Fanzhang Li}
\affiliation{%
  \institution{School of Computer Science and Technology, Soochow University}
  \country{China}
}
\email{email: lfzh@suda.edu.cn}
%\footnote{$^*$Corresponding authors}
%\footnote{Corresponding author}

%%
%% By default, the full list of authors will be used in the page
%% headers. Often, this list is too long, and will overlap
%% other information printed in the page headers. This command allows
%% the author to define a more concise list
%% of authors' names for this purpose.
\renewcommand{\shortauthors}{Zhu, Fan, Qian, and Li}
%%
%% The abstract is a short summary of the work to be presented in the
%% article.
\begin{abstract}
Video Object Segmentation (VOS) aims to accurately segment target objects from consecutive video frames and track the changes of the objects in each frame of the video. Conventional VOS methods typically demand substantial quantities of pixel-level labeled video sequences for fully supervised learning, which limits the performance of the model in sparse video scenes, while existing VOS methods have limited adaptability to global changes in objects. Based on this observation, in this paper, we propose self-supervised VOS with Sliding window, Wavelet transform and optimal Transport (SWT), a self-supervised VOS framework entirely trained on static dataset using contrastive learning. Firstly, a rolling sample buffer reuses overlapping groups of independently sampled images across successive updates. Secondly, to address the long-distance modeling difficulty caused by simple convolutional structures, we introduce wavelet transform to expand the receptive field of convolutional kernels, thus improving the model's representational capability. Finally, we incorporate optimal transport to assist the model in finding the globally optimal match between the target across two frames, improving the model's ability to handle non-rigid deformations of objects. SWT only requires training on the COCO dataset once and achieves excellent results on five VOS datasets as well as an additional body part propagation dataset. The code will be released soon at \url{https://github.com/machine928/SWT.git}.
\end{abstract}

%% Candidate CCS term; confirm the final selection in the ACM system.
\ccsdesc[500]{Computing methodologies~Computer vision problems}

%%
%% Keywords. The author(s) should pick words that accurately describe
%% the work being presented. Separate the keywords with commas.
\keywords{Self-Supervised, Video Object Segmentation, Wavelet Transform, Optimal Transport}

%%
%% This command processes the author and affiliation and title
%% information and builds the first part of the formatted document.
\maketitle

\section{Introduction}
\label{sec:intro}

    Recent visual learning systems support efficient multimodal in-context learning~\cite{gao2025aim}, unified text--vision reasoning~\cite{qin2025unicot}, and joint video generation and understanding~\cite{qin2026univigu}.

    Video Object Segmentation (VOS) identifies and tracks video objects using pixel-level masks. Self-supervised VOS reduces reliance on labor-intensive annotations by learning from unlabeled data. Visual correspondence learning establishes cross-frame associations without human supervision\cite{jabri2020space,lai2020mast,xu2020self,li2022locality,son2022contrastive}, exploiting video continuity through feature alignment and motion patterns. It supports downstream tasks including optical flow estimation\cite{yin2018geonet}, video object tracking\cite{cheng2023segment,cao2023observation}, and VOS\cite{miles2023mobilevos,vondrick2018tracking}.

    \begin{figure}[t]
        \centering
        \includegraphics[width=0.90\linewidth]{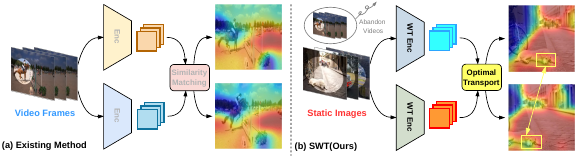}
        \caption{
           Comparison with current self-supervised VOS approaches. Based on the contrastive learning paradigm, existing methods (a) typically rely on unlabeled video data for training, while our method (b) only requires randomly cropped image pairs from static images as training samples. Compared to existing methods, our Wavelet Transform (WT) encoder achieves a larger effective receptive field, and we employ optimal transport mechanism to learn cross-frame object correspondence, which improves our method a stronger long-distance modeling capability. The yellow box in (b) represents that our method is able to form a better cross-frame correspondence of objects.
        }
        \label{fig:pipeline_simp}
    \end{figure}
 
    Visual correspondence learning aligns positive sample pairs and separates negative ones in latent space\cite{chen2020simple}, requiring many training pairs to learn discriminative representations. Video acquisition costs can limit the available samples. Convolutional architectures used by self-supervised VOS methods\cite{lu2020learning,lai2020mast,xie2021propagate} extract local features effectively but have limited long-range receptive fields, making global spatio-temporal consistency difficult to capture. Sample scarcity and limited global matching further complicate segmentation of rapidly moving or deforming objects.

    SWT addresses sample utilization and long-range modeling jointly. A rolling sample buffer reuses overlapping groups of independently sampled images across successive updates. Cascaded wavelet decomposition\cite{finder2024wavelet} and frequency-specific small-kernel convolutions emphasize low-frequency components and enlarge the effective receptive field. To handle object displacement and deformation, optimal transport\cite{zhao2021towards} with differentiable Sinkhorn\cite{cuturi2013sinkhorn} iterations computes globally optimal matching plans. Formulating correspondence as mass transport supports smooth, consistent matching under spatial distortions or occlusions. Figure \ref{fig:pipeline_simp} illustrates the resulting cross-frame correspondence.

    The key design insight of SWT is to align these components with the static-image self-supervised correspondence pipeline rather than treating them as independent add-ons. The rolling buffer improves sample reuse, wavelet decomposition makes long-range context within static crops more comparable, and optimal transport converts local pairwise similarities into a globally constrained matching plan. Together, they provide correspondence supervision for VOS without requiring video annotations during training.

    Trained solely on static images, SWT improves \(\mathcal{J} \& \mathcal{F}\) over the baseline by 1.6\% on DAVIS16-val\cite{perazzi2016benchmark} and 1.3\% on DAVIS17-val\cite{pont20172017}. Our main contributions are:

    \begin{itemize}
      \item \textbf{Novel Training Strategy.} We use a rolling sample buffer to reuse overlapping sample groups across successive optimization steps while preserving computational efficiency.
    
      \item \textbf{Wavelet Transform Encoder.} To overcome the limitations of convolutional architectures in long-range modeling, we introduce wavelet transform decomposition to expand the effective receptive field, improving the model's capacity to learn distant relationships with moderate computational overhead

      \item \textbf{Optimal Transport for Global Matching.} To address the global matching challenge of objects undergoing displacement and deformation in videos, we integrate optimal transport theory to achieve globally optimal object correspondence across frames.

      \item \textbf{Advanced experimental results.} We evaluate our approach on several datasets including DAVIS-16\allowbreak\cite{perazzi2016benchmark}, DAVIS-17\allowbreak\cite{pont20172017}, YouTubeVOS-18\allowbreak\cite{xu2018youtube}, YoutubeVOS-19\allowbreak\cite{yang2019video} and VIP\allowbreak\cite{zhou2018adaptive}. Quantitative results show that the proposed method significantly outperforms existing advanced methods.
    \end{itemize}

\section{Related Work}
\label{sec:relate_work}
    \textbf{Self-supervised VOS} is a VOS learning paradigm that requires no annotations during the training phase. Self-supervised learning tasks learn visual representations by constructing proxy tasks, and many contrastive learning-based methods\cite{wang2021dense, xie2021detco, yang2021instance} have been successfully implemented for various image-level downstream tasks. The core idea is to pull feature embeddings of the same class closer while pushing those of different classes apart. Some research learns video representations in a self-supervised manner by leveraging visual correspondences: CorrFlow\cite{lai2019self} and MAST\cite{lai2020mast} learn representations by reconstructing color frame structures. TimeCycle\cite{wang2019learning} and CRW\cite{jabri2020space} learn representations that reflect correspondences from temporal cycle consistency. VFS\cite{xu2021rethinking} learns correspondences by simply using video frame-level similarity. PixPro\cite{xie2021propagate} leverages pixel-level consistency across different views of the same image. VideoHiGraph\cite{qin2023exposing} proposes a complementary approach that utilizes video supervision from a graph and node-level perspective, generating structural representations through learnable hidden graphs to further explore more flexible supervision. HVC\cite{pei2024dynamic} injects dynamic signals into images, embedding static and dynamic consistency into the original image-level visual correspondence learning. These methods demonstrate that self-supervised learning can be elegantly extended to VOS scenarios. 
    
    \textbf{Temporal consistency.} Temporal consistency is also studied in video synthesis. MuseTalk~\cite{zhang2024musetalk} uses spatiotemporal sampling for identity-preserving video dubbing, while AnyTalker~\cite{zhong2025anytalker} uses identity-aware attention for multi-person generation. SoulX-FlashTalk~\cite{shen2025soulx} and SoulX-FlashHead~\cite{yu2026soulx} combine bidirectional distillation with mechanisms for stable streaming generation. In VOS, temporal consistency concerns persistent object identities and accurate mask propagation.

    \textbf{Wavelet transform} is an efficient tool for signal analysis and processing. Recently, wavelet neural networks (WNNs) have achieved remarkable success and been extensively used in various tasks. Duan et al.\cite{duan2015semg} compared the classification accuracy of WNNs and conventional neural networks on surface electromyography signals, as well as the selection of different mother wavelets. Ji et al.\cite{ji2019research} employ the Monte Carlo method with Gaussian wavelet-type activation functions to design neural networks and apply them to image classification. More pertinently, WCC\cite{finder2022wavelet} employs wavelets to construct feature maps for improving CNN efficiency. WIRE\cite{saragadam2023wire} utilizes wavelet basis functions as nonlinear activations within its implicit neural representation framework. WTConv\cite{finder2024wavelet} utilizes cascaded wavelet transforms to expand the effective receptive field at the cost of only minimal additional trainable parameters. CFWD\cite{cfwd} propose a novel and robust low-light image enhancement method via CLIP-Fourier guided wavelet diffusion. These works fully demonstrate the great contribution of wavelet transform in image processing.
    
    \textbf{Optimal transport} is a classical theoretical framework in mathematics that studies how to transform one probability distribution into another with minimal cost. It has gained significant attention in machine learning and computer vision. To accelerate convergence and efficiently handle large-scale problems, Cuturi et al.\cite{cuturi2013sinkhorn} introduce the Sinkhorn algorithm\cite{sinkhorn1967diagonal} for computing approximate transport couplings with entropy regularization. Due to its exceptional capability in distribution matching, optimal transport has been applied to various theoretical and practical tasks: OTFace\cite{qian2022otface} employs optimal transport to characterize the distributional differences of high-level convolutional features for improving face representation learning. UniOT\cite{chang2022unified} designs an adaptive partial alignment method based on optimal transport for domain matching. CSOT\cite{chang2023csot} proposes a novel optimal transport called curriculum and structure-aware optimal transport. DIML\cite{zhao2023diml} decomposes the similarity between two images into multiple partial similarities and their contributions to the overall similarity using optimal matching. AWT\cite{zhu2024awt} leverages optimal transport to mine semantic associations in visual-language spaces. SD-MOT\cite{sd_mot} defines a semi-discrete multi-view optimal transport method for the application of incomplete multi-view clustering (IMC) in the Internet of Things. Matching stability also arises in supervised text spotting, where DNTextSpotter~\cite{qiao2024dntextspotter} uses denoising training to address unstable bipartite assignments. SWT formulates cross-view feature matching as an entropy-regularized transport problem.
\begin{figure*}[!t]
  \centering
    \includegraphics[width=0.94\linewidth]{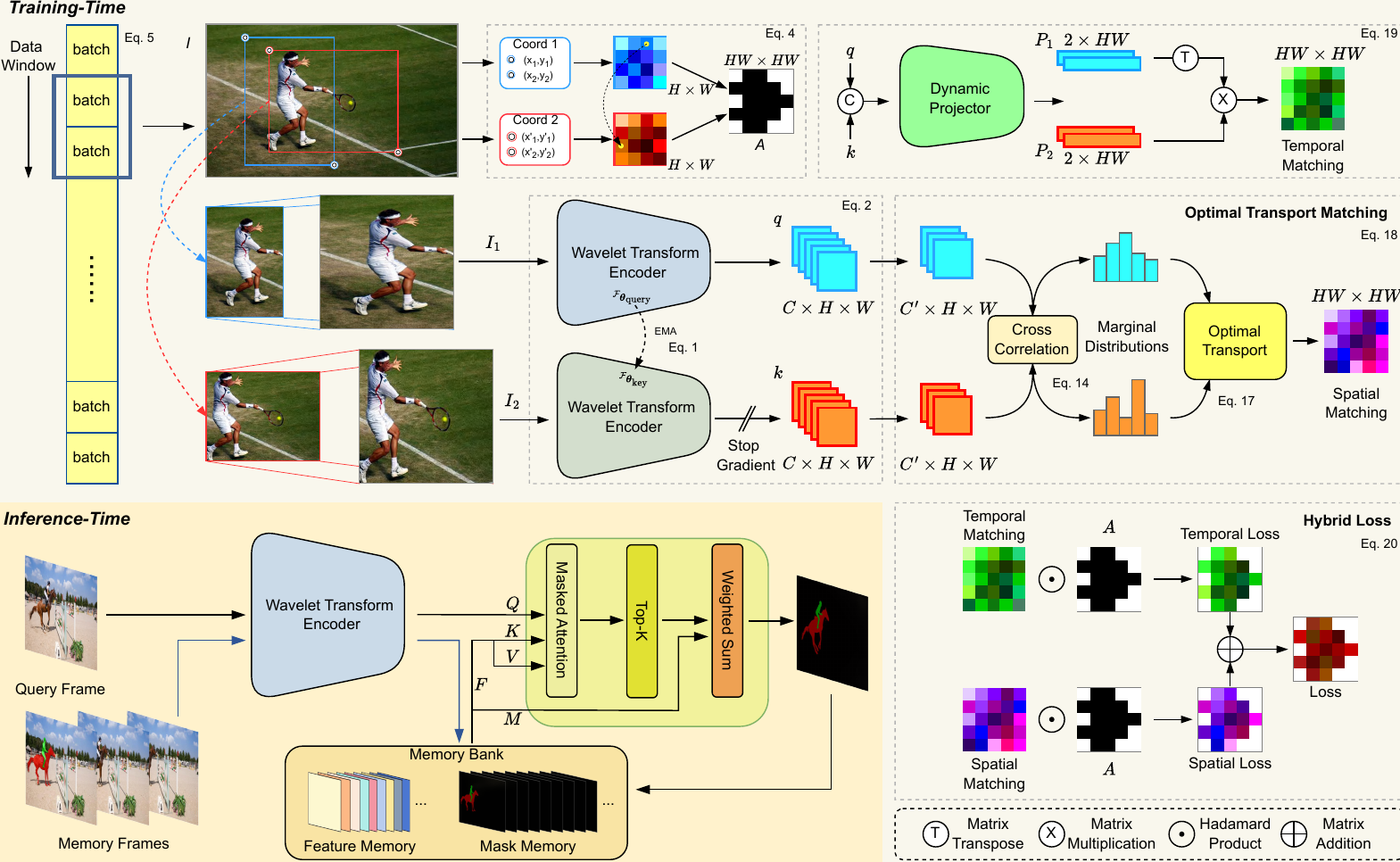}
    \caption{
        The pipeline of our proposed method. \textbf{Training-time:} Independently sampled static-image mini-batches are organized by the rolling buffer described in Section \ref{subsec:window}. Two images  \(I_1,I_2 \in \mathbb{R} ^ {3 \times h \times w}\) are then cropped on the static image with a random aspect ratio of 3/4 to 4/3, resize them, and use them as model inputs. These two images are used to compute a similarity matrix \(A \in \mathbb{R} ^ {HW \times HW}\), where elements with similarity below a set threshold are set to 0 and others to 1. \(I_1\) and \(I_2\) are fed into query encoder and key encoder respectively to obtain features \(q,k \in \mathbb{R} ^ {3 \times H \times W}\). After grouping and fusing \(q\) and \(k\), we perform cross correlation to generate marginal distributions, then use optimal transport to minimize the transport cost as spatial matching \(L_{spatial}\). Next, \(q\) and \(k\) are concatenated in different orders and fed into a dynamic projector to generate two pseudo-dynamic representations \(P_1,P_2 \in \mathbb{R} ^ {2 \times HW}\), which are used to compute temporal matching \(L_{temporal}\). \(L_{spatial}\) and \(L_{temporal}\) are each multiplied by matrix \(A\) to produce spatial loss \(\mathcal{L}_{spatial}\) and temporal loss \(\mathcal{L}_{temporal}\), which are then summed as the final loss \(\mathcal{L}_{hybrid}\). \textbf{Inference-time:} Given the initial frame's mask \(M_0\), the initial frame is encoded by query encoder and stored in memory bank along with \(M_0\). For subsequent query frames, after encoding, masked attention is performed between the query features and memory features to retrieve the top K most similar memory masks. These masks are then weighted and fused to predict the query frame's mask, which is finally stored in memory.
    }
    \label{fig:main}
\end{figure*}

\section{Approach}
\label{sec:approach}
    As illustrated in Figure \ref{fig:main}, our method adopts a self-supervised training and semi-supervised inference pipeline. Specifically, during the training phase, the model is trained in a completely human-annotation-free manner. In the inference phase, given a video sequence \(I=\left\{I_0, I_1, \ldots, I_t\right\}\) and the object mask \(M_0\) of the first frame, the mask is propagated by leveraging inter-frame similarities. We provide more architectural details in subsequent sections of this chapter.

%===================================Method Overview===================================
\subsection{Method Overview}
\label{subsec:overview}
    In this study, we apply visual correspondence learning to self-supervised VOS, with feature representations learned entirely from unannotated static images. To address the long-range context matching problem in visual correspondence, we innovatively introduce wavelet transform and optimal transport. Our approach is summarized below.

    As illustrated in the Figure \ref{fig:main}, the training mini-batches are organized by the rolling sample buffer described in Section \ref{subsec:window}. Given a raw static image \(I \in \mathbb{R}^{3 \times H_s \times W_s}\), we randomly crop two frames \(I_{1,} I_2 \in \mathbb{R}^{3 \times h \times w}\) from the static image with aspect ratios varying randomly between 3/4 and 4/3\cite{pei2024dynamic}. \(I_1\) and \(I_2\) are then fed into the query encoder \(\mathcal{F}_{\theta_{\text {query}}}\) and key encoder \(\mathcal{F}_{\theta_{\text {key}}}\) respectively. Both encoders consist of a backbone and a projection head. To establish an asymmetric structure between the query and key encoders, we incorporate an additional prediction head in the query encoder following common practice\cite{xie2021self}. According to\cite{caron2021emerging}, the query encoder parameters \(\theta_{\text{query}}\) are updated via gradients, while the key encoder parameters \(\theta_{\text{key}}\) are updated follow the momentum update rule: 
    \begin{equation} \boldsymbol{\theta}_{\text {key}} \leftarrow m 
        \boldsymbol{\theta}_{\text {key}}+(1-m) 
        \boldsymbol{\theta}_{\text {query}}, m \in[0,1],
        \label{eq:ema}
    \end{equation}
    where \(m\) is the momentum coefficient. The feature maps of two differently cropped views of the image are computed by \(\mathcal{F}_{\theta_{\text {query}}}\) and \(\mathcal{F}_{\theta_{\text {key}}}\) respectively, followed by \(L_2\) normalization to obtain:
    \begin{equation}
    \boldsymbol{q}, \boldsymbol{k}=l_2\left(\mathcal{F}_{\boldsymbol{\theta}_{\text {query }}}\left(\boldsymbol{I}_1\right)\right), l_2\left(\mathcal{F}_{\boldsymbol{\theta}_{\text {key}}}\left(\boldsymbol{I}_2\right)\right) \in \mathbb{R}^{C \times H \times W} .
    \end{equation}
    We apply optimal transport to the obtained \(q\) and \(k\) to establish spatial matching (more details in Section \ref{subsec:transport}). Simultaneously, we concatenate \(q\) and \(k\) in different orders and feed them into the dynamic projector to generate forward and backward pseudo-dynamics for calculating  temporal matching.

    Below we describe the selection of positive regions. Given two cropped views of different sizes from an image, resizing causes feature position offsets. We randomly crop the image to obtain two views while recording their original coordinates (top-left and bottom-right corners). Based on the spatial positions of the two cropped views in the original image, we generate the transformed pixel coordinates of the views as:
    \begin{equation}
        \begin{aligned}
        & \boldsymbol{x}_1^o, \boldsymbol{y}_1^o=\mathcal{T}\left(\boldsymbol{x}_1, \boldsymbol{y}_1\right), \\
        & \boldsymbol{x}_2^o, \boldsymbol{y}_2^o=\mathcal{T}\left(\boldsymbol{x}_2, \boldsymbol{y}_2\right),
        \end{aligned}
        \label{eq:wrapping}
    \end{equation}
    where \(\mathcal{T}\left(\boldsymbol{.}, \boldsymbol{.}\right)\) denotes the warping operation, and \(x^o, y^o \in \mathbb{R}^{H \times W}\) represent the transformed horizontal and vertical pixel coordinates respectively.

    Since the original coordinates of these views are known, we use the warping operation in Eq. \ref{eq:wrapping} to transform the projected features back to the original image space. Subsequently, we compute the Euclidean distance \(\mathcal{D}\) between the two sets of coordinates of the feature maps in the original image space. This distance measures feature similarity and helps the model establish visual correspondences. To define the spatial neighborhood in the local feature space, we set a positive radius \(r\) for feature vectors in the overlapping regions of the two cropped views. Finally, the positive region selection mask \(A \in \mathbb{R}^{HW \times HW}\) for the two views can be expressed as:
    \begin{equation}
        \boldsymbol{A}(i, j)= \begin{cases}
        1, & \text { if } \mathcal{D}(i, j) \leq r \\
        0, & \text { otherwise }\end{cases},
        \label{eq:pos}
    \end{equation}
    where \(i, j\) represent pixel coordinates in the two cropped views. We set the positive radius \(r\) to 0.1, and use \(A\) as our self-supervised signal for subsequent training.

%===================================Sliding Window===================================
\subsection{Sliding-Window Sample Buffer}
\label{subsec:window}
    In SWT, the sliding window is implemented as a rolling buffer over the mini-batch stream. Let \(B=\{b_i\}_{i=1}^{n}\) denote the batches produced by the epoch sampler, where \(b_i \in \mathbb{R}^N\) is the \(i^{th}\) mini-batch and \(N\) is both the mini-batch size and the number of samples refreshed at each step. The buffer holds \(c\) mini-batches, and the sample group used at step \(i\) is
    \begin{equation}
        W_i=\left\{b_{1+i}\left|b_{2+i}\right| b_{3+i}|\ldots| b_{c+i}\right\},
        i \in[0, n-c],
        \label{eq:set}
    \end{equation}
    where .\(|\). denotes concatenation along the first dimension and \(W_i \in \mathbb{R}^{cN}\). Once the buffer is full, each update replaces the oldest mini-batch with the incoming one. Adjacent buffers \(W_i\) and \(W_{i+1}\) therefore share \((c-1)N\) samples, allowing each image to participate in up to \(c\) consecutive updates. We use a buffer length of \(c=4\) with \(N=16\), yielding 64 samples per update: 48 are retained from the preceding step and 16 are newly added. The buffer is cleared after each epoch. Since COCO images are sampled independently, this schedule changes only their reuse across optimization steps and introduces no temporal or spatial supervision between images. Additional experimental details are provided in Section \ref{sec:experiments}.

%===================================Wavelet Transform===================================
\subsection{Wavelet Transform}
\label{subsec:wavelet}
    In this study, we employ an efficient and concise Haar Wavelet Transform\cite{finder2022wavelet}. Given an input \(I\in\mathbb{R}^{C \times h \times w}\), for \(2D\) Haar WT, we implement four distinct filters through depth-wise convolutions (stride=2):
    \begin{equation}
        \begin{aligned}
        & f_{L L}=\frac{1}{2}\left[\begin{array}{ll}
        1 & 1 \\
        1 & 1
        \end{array}\right], f_{L H}=\frac{1}{2}\left[\begin{array}{ll}
        1 & -1 \\
        1 & -1
        \end{array}\right], \\
        & f_{H L}=\frac{1}{2}\left[\begin{array}{cc}
        1 & 1 \\
        -1 & -1
        \end{array}\right], f_{H H}=\frac{1}{2}\left[\begin{array}{cc}
        1 & -1 \\
        -1 & 1
        \end{array}\right],
        \label{eq:filter}
        \end{aligned}
    \end{equation}
    where \(f_{LL}\) denotes a low-pass filter, with \(f_{LH}\),\(f_{HL}\),\(f_{HH}\) constituting the high-pass filter. For each input channel, the convolution output is given by:
    \begin{equation}
        \begin{aligned}
            \left[I_{L L}, I_{L H}, I_{H L}, I_{H H}\right]=\operatorname{Conv}\left(  
            \left[f_{L L}, f_{L H}, f_{H L}, f_{H H}\right], I\right),
        \end{aligned}
        \label{eq:single_wt}
    \end{equation}
    it has four channels, each with spatial dimensions half of I. \(I_{L L} \in \mathbb{R}^{C \times \frac{h}{2} \times \frac{w}{2}}\) is the low-pass component of I, while \(I_{L H},I_{H L},I_{H H} \in \mathbb{R}^{C \times \frac{h}{2} \times \frac{w}{2}}\) represent horizontal, vertical, and diagonal high-frequency components of \(I\), respectively. Given the orthonormality of the filters in Eq. \ref{eq:filter}, the inverse wavelet transform (IWT) can be achieved through transposed convolution operations:
    \begin{equation}
        \begin{aligned}
        I=\text { Conv-transposed }( & {\left[f_{L L}, f_{L H}, f_{H L}, f_{H H}\right] } \\
        & {\left.\left[I_{L L}, I_{L H}, I_{H L}, I_{H H}\right]\right) }.
        \end{aligned}
    \end{equation}
    Then, a cascaded wavelet decomposition is achieved by recursively decomposing the low-frequency components. The decomposition formula at each level is as follows:
    \begin{equation}
        I_{L L}^{(i)}, I_{L H}^{(i)}, I_{H L}^{(i)}, I_{H H}^{(i)}=
        \mathrm{WT}\left(I_{L L}^{(i-1)}\right),
        \label{eq:cscade_wt}
    \end{equation}
    where \(i\) is the current level.

\begin{figure}[t]
    \centering
    \includegraphics[width=0.85\linewidth]{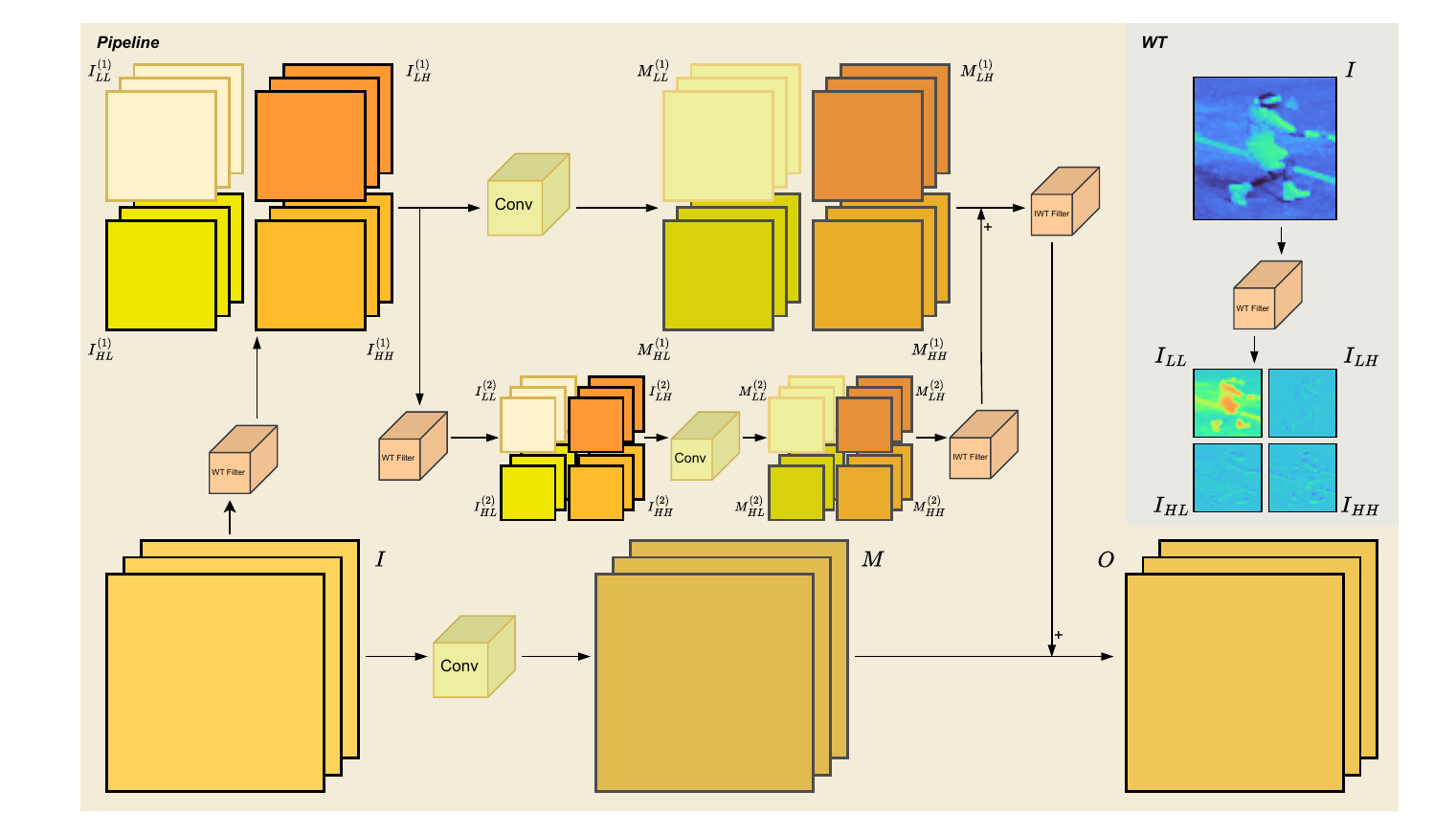}
    \caption{
      The pipeline of wavelet convolution. The architecture employs a cascaded wavelet transform framework. For the input feature map, we first apply a 2D wavelet transform to decompose the image into four channels: low-frequency, horizontal, vertical, and diagonal high-frequency. Convolution operations are then performed on these four channels in the wavelet domain. The results are reconstructed via inverse wavelet transform and used as residuals, which are combined with the original convolved input through residual connections to produce the final output. Notably, during the second-level wavelet transform, we only process the low-frequency channel.    
    }
   \label{fig:wtconv}
\end{figure}

    In our method, the following steps are performed. First, the WT is used to filter and downsample the input's low-frequency and high-frequency content. Then, small-kernel depth-wise convolutions are applied to each subband before reconstruction through the IWT. In other words, the entire process proceeds as follows:
    \begin{equation}
        O=\operatorname{IWT}(\operatorname{Conv}(W, \mathrm{WT}(I))),
    \end{equation}
    where \(I\in\mathbb{R}^{C \times h \times w}\) is the input, \(O\in\mathbb{R}^{C \times h \times w}\) is the output after IWT, and \(W\) denotes the weight tensor of the depth-wise convolutional kernel with an input channel count four times that of \(I\). This operation not only decouples convolution across frequency components but also enables smaller kernels to operate over larger receptive fields of the original input, effectively increasing their perceptual scope. Leveraging Eq. \ref{eq:cscade_wt}, we further enhance the current level-1 combination:
    \begin{equation}
        \begin{aligned}
        I_{L L}^{(i)}, I_H^{(i)} & =\mathrm{WT}\left(I_{L L}^{(i-1)}\right) \\
        M_{L L}^{(i)}, M_H^{(i)} & =\operatorname{Conv}\left(W^{(i)},\left(I_{L L}^{(i)}, I_H^{(i)}\right)\right),
        \end{aligned}
        \label{eq:cscade_wt_conv}
    \end{equation}
    where \(I_{L L} ^ {(i)} \in \mathbb{R}^{C \times \frac{h}{2 ^ i} \times \frac{w}{2 ^ i}}\) represents the low-frequency map at the \(i ^ {th}\) level, and \(I_{H} ^ {(i)} \in \mathbb{R}^{C \times \frac{h}{2 ^ i} \times \frac{w}{2 ^ i}}\) denotes the three high-frequency maps at the \(i ^{th}\) level. \(M_{L L}^{(i)}, M_H^{(i)} \in \mathbb{R}^{C \times \frac{h}{2 ^ i} \times \frac{w}{2 ^ i}}\) are the results after performing convolution on \(I_{L L} ^ {(i)},I_{H} ^ {(i)}\). To combine outputs from different frequencies, the final output can be expressed as:
    \begin{equation}
        O^{(i)}=\operatorname{IWT}\left(M_{L L}^{(i)}+O^{(i+1)}, M_H^{(i)}\right),
    \end{equation}
    where \(O ^ {(i)}\in\mathbb{R}^{C \times \frac{h}{2 ^ i} \times \frac{w}{2 ^ i}}\) is the cumulative output after the \(i ^ {th}\) level. Figure \ref{fig:wtconv} illustrates the above process. Furthermore, when applying wavelet convolution to the encoder, we utilize \(1 \times 1\) convolutions to transform the spatial dimensions and channels of the feature maps. 

%===================================Optimal Transport===================================
\subsection{Optimal Transport}
\label{subsec:transport}
    We employ optimal transport theory for spatial matching, which determines the optimal transport mapping relation by finding the solution with the lowest transport cost between two probability distributions, subject to marginal constraints. Consider a source distribution \(\mu_q\) defined on probability spaces \(\mathcal{U}\) and a target distribution \(\mu_k\) defined on probability spaces \(\mathcal{V}\), the optimal transport plan can be derived by computing the minimal Wasserstein metric between the source and target distributions:
    \begin{equation}
        \pi^*=\underset{\pi \in \Pi\left(\mu_q, \mu_k\right)}{\arg \inf }
        \int_{\mathcal{U} \times \mathcal{V}} c(u, v) \mathrm{d} \pi(u, v),
        \label{eq:ot}
    \end{equation}
    where \(\pi^*\) denotes the optimal transport plan, \(\Pi\left(\mu_q, \mu_k\right)\) represents the joint probability distribution between \(\mu_q\) and \(\mu_k\), and \(c(u,v)\) stands for the cost function associated with transport.

    We start with two feature maps \(q, k \in \mathbb{R}^{C \times H \times W}\) obtained by a backbone. First, we perform group-wise fusion along the channel dimension and then flatten them to obtain \(z_q, z_k \in \mathbb{R}^{C^{\prime} \times H W}\). To make this step reproducible, we split the channel dimension into contiguous groups using the same fixed grouping rule for \(q\) and \(k\). In our implementation, the projection output dimension is \(C=256\), and the channel-group size is set to 32, yielding \(G=C/32=8\) groups. For each spatial location, the features within each group are summed into one grouped response, and the resulting grouped maps are \(\ell_2\)-normalized and flattened to \(z_q,z_k \in \mathbb{R}^{G \times HW}\), i.e., \(C^{\prime}=G=8\). This fusion step contains no additional learnable layer; the grouped features are then used for cross-correlation marginal initialization and OT cost computation. Simultaneously, we initialize the marginal distributions \(\mu_q\), \(\mu_k\) using cross correlation:
    \begin{equation}
        \begin{aligned}
            & \mu_q=\mathcal{M}(z_q)^{\top} \times z_k \in R^{H W}, \\
            & \mu_k=\mathcal{M}(z_k)^{\top} \times z_q \in R^{H W}.
        \end{aligned}
        \label{eq:cc}
    \end{equation}
    Where \(\mathcal{M}(.)\) represents the calculation of the mean for the second dimension. We also attempt to initialize \(\mu_q\) and \(\mu_k\) using uniform distributions:
    \begin{equation}
        \mu_q^{i}=\mu_k^{i}=\frac{1}{H W}, i \in [1,HW],
        \label{eq:uniform}
    \end{equation}
    which indicates that each position similarity carries the same weight in the overall similarity measure. The marginal distributions ought to reflect the relative significance of every spatial position. Simply using a uniform distribution implies that we aim to treat all features with identical weighting, which may not be desirable in certain scenarios. Experimental results comparing these two initialization approaches are presented in Section \ref{sec:experiments}.

    The transport cost per unit between i and j is mathematically expressed as:
    \begin{equation}
        C_{i, j}=c(i, j)=d\left(z_q^{i}, z_k^{j}\right), i\ j \in [1,HW],
    \end{equation}
    the transport cost function \(c\) is defined using the Euclidean distance between the \(\ell_2\)-normalized grouped features. In this discrete case, the transport plan \(\pi\) that aligns the two distributions similarly adopts a discrete form. For given discrete distributions \(\mu_q\) and \(\mu_k\), the classical optimal transport formulation can be written as:
    \begin{equation}
        \begin{aligned}
            & T^*=\underset{T \geq 0}{\arg \min } \operatorname{tr}\left(C T^{\top}\right) \\
            & \text { subject to } \quad T \mathbf{1}=\mu_q, \quad T^{\top} \mathbf{1}=\mu_k.
        \end{aligned}
        \label{eq:sink}
    \end{equation}
     The optimal matching flow between these two distributions is represented by the matrix \(T^* \in \mathbb{R}^{HW \times HW}\), which can also be seen as the structural matching scheme between the two views. We employ the Sinkhorn divergence algorithm\cite{cuturi2013sinkhorn} to implement this process. Unless otherwise stated, we use cross-correlation marginals rather than uniform marginals. The entropic regularization coefficient is set to 0.05, the maximum number of Sinkhorn iterations is 100, the early-stopping threshold is \(10^{-1}\), and the numerical stabilizer is \(10^{-8}\).

    We have demonstrated how to compute the optimal transport between two distributions. Next, we will explain how to use optimal transport for model training. Having obtained the optimal matching flow \(T^*\) through Eq. \ref{eq:sink}, we first compute the Euclidean distance \(\mathcal{D}\) between \(z_q\) and \(z_k\). We then perform a dot product of \(T^*\) and \(\mathcal{D}\), and add the result to \(\mathcal{D}\) in a residual manner as the spatial matching output. This process is formally expressed as:
    \begin{equation}
        L_{spatial}=\mathcal{D}\left(z_q, z_k\right) \cdot T^*+\mathcal{D}\left(z_q, z_k\right).
        \label{eq:spatial_logit}
    \end{equation}
    \(L_{spatial} \in \mathbb{R}^{HW \times HW}\) represents the spatial matching output, which will be used for subsequent hybrid loss computation.

    Compared with standard attention or transformer-style global matching, the optimal transport formulation is more suitable for our static-image self-supervised correspondence setting. Attention mainly aggregates pairwise similarities and normalizes them locally, whereas optimal transport imposes marginal constraints on the matching plan and therefore encourages a globally balanced soft assignment between the two cropped views. This mass-conservation constraint reduces degenerate many-to-one matches and provides a structured training signal without requiring additional video-level supervision. Moreover, in SWT, optimal transport is used as a training-time matching loss; it does not introduce an extra inference-stage matching module or change the memory-based propagation pipeline.

\subsection{Details}
\label{subsec:details}
    \textbf{Hybrid Loss.} Our model employs a hybrid spatio-temporal loss during training. Specifically, given the feature maps \(q,k \in \mathbb{R}^{C \times H \times W}\) extracted from the backbone, we concatenate \(q\) and \(k\) along the channel dimension in different orders. The dynamic projector \(\mathcal{F}_{pseudo}\), implemented as the Flow module, takes the concatenated feature \([q,k]\in\mathbb{R}^{2C\times H\times W}\) as input. With the default \(C=256\), it consists of \(1\times1\) Conv\(512\rightarrow256\), BatchNorm2d, ReLU, \(1\times1\) Conv\(256\rightarrow2\), and BatchNorm2d. The same projector is applied to the two input orders to produce \(P_1,P_2 \in \mathbb{R}^{2 \times HW}\), which are normalized along the channel dimension before temporal matching. We calculate their cosine similarity to obtain the temporal matching:
    \begin{equation}
        \begin{aligned}
            & P_1=\mathcal{F}_{\text {pseudo }}(q, k), P_2=\mathcal{F}_{\text {pseudo }}(k, q), \\
            & L_{temporal}=\frac{\left\langle P_1, P_2\right\rangle}{\left|P_1 \| P_2\right|} \in R^{H W \times H W}.
        \end{aligned}
    \end{equation}
    Building on this, we combine the temporal matching score \(L_{temporal}\) with the previously obtained spatial matching score \(L_{spatial}\) (Eq. \ref{eq:spatial_logit}). Each is multiplied with the affinity matrix \(A\) (Eq. \ref{eq:pos}) via dot product to compute the temporal loss \(\mathcal{L}_{temporal}\) and spatial loss \(\mathcal{L}_{spatial}\). The hybrid loss \(\mathcal{L}_{hybrid}\) is then derived by summing these two terms:
    \begin{equation}
        \begin{aligned}
            & \mathcal{L}_{\text {temporal }}=-\frac{\sum_{i=1}^{H W} \sum_{j=1}^{H W} 
            (L_{\text {temporal }}^{i, j} \cdot A^{i, j})}
            {\sum_{i=1}^{H W} \sum_{j=1}^{H W} A^{i, j}}, \\
            & \mathcal{L}_{\text {spatial }}=-\frac{\sum_{i=1}^{H W} \sum_{j=1}^{H W}
            (L_{\text {spatial }}^{i, j} \cdot A^{i, j})} 
            {\sum_{i=1}^{H W} \sum_{j=1}^{H W} A^{i, j}}, \\
            & \mathcal{L}_{\text {hybrid }}=\alpha \mathcal{L}_{\text {temporal }} + \mathcal{L}_{\text {spatial }},
        \end{aligned}
    \end{equation}
    where \(\alpha = 1\) assigns equal weight to the temporal and spatial losses. This default is selected based on the sensitivity analysis in Table \ref{tab:ablation_sensitivity}(d). This hybrid loss mechanism effectively incorporates spatio-temporal information to provide a more comprehensive supervised signal for model training.

    \textbf{Training.} We use COCO\cite{lin2014microsoft} as the static training dataset and train our model on a single NVIDIA A100 GPU. To generate input pairs, we apply random cropping to produce two images from each original image, resizing all crops to \(256 \times 256\) pixels. The entire model is implemented in PyTorch, with a window size set to 64 for training over 20 epochs. The Adam optimization method is adopted, configured with an initial learning rate of \(10^{-3}\) and 0 weight decay.

\begin{table}[t]
    \centering
    \small
    \caption{Implementation hyperparameters used for the reported SWT model.}
    \label{tab:implementation_hyperparameters}
    \begin{tabular}{p{0.38\linewidth}p{0.54\linewidth}}
        \hline
        \textbf{Item} & \textbf{Setting} \\ \hline
        Static training data & COCO train split; two random crops per image \\
        Input augmentation & Random resized crop; area scale \([0.0,1.0]\); aspect ratio \([3/4,4/3]\); resized to \(256 \times 256\) \\
        Normalization & Mean \((0.4914,0.4822,0.4465)\); std. \((0.2023,0.1994,0.2010)\) \\
        Backbone and projection & ResNet-18; strides removed from layer3/layer4; projection hidden/output dimension 256; prediction hidden dimension 256 \\
        Key encoder update & Momentum coefficient initialized as 0.99 and increased to 1 with a cosine schedule \\
        Optimization & Adam; learning rate \(10^{-3}\); weight decay 0; 20 epochs; seed 42; mixed precision enabled \\
        Sample buffer & Buffer length \(c=4\); mini-batch/refresh size \(N=16\); capacity \(cN=64\); cleared after each epoch \\
        Positive mask and loss & Positive radius \(r=0.1\); temporal-loss weight \(\alpha=1\) \\
        Optimal transport & Channel-group size 32; Euclidean cost on \(\ell_2\)-normalized grouped features; cross-correlation marginal initialization; entropic coefficient 0.05; maximum Sinkhorn iterations 100 \\
        Inference & Top-\(K=15\); temperature 0.1; local radius 12; context size 50; original input resolution retained \\
        \hline
    \end{tabular}
\end{table}

    \textbf{Inference.} Following the approach of CRW\cite{jabri2020space}, we propagate masks by leveraging the similarity between the current frame and previous frames. Similar to\cite{hu2022semantic}, we adopt MoCo\cite{he2020momentum} as the baseline model to extract rich semantic representations. Specifically, given a video sequence \(I=\left\{I_0, I_1, \ldots, I_t\right\}\) and the initial frame mask \(M_0\), we encode the first frame features \(F_0\), then send \(F_0\) and \(M_0\) to a memory bank. For each subsequent query frame, its encoded features interact with the memory frames via masked attention\cite{cheng2022masked}, retrieving the top-K most similar memory frames. The corresponding masks of these frames are then weighted and fused to predict the query frame mask. The predicted mask is subsequently added to the memory bank, and this process repeats until the entire video sequence is segmented.

\begin{table}[!t]
\setlength{\tabcolsep}{6.8pt}
\centering
\caption{Quantitative comparison to methods on DAVIS17-val. The three highest-scores are highlighted in colored text: \textcolor{red}{red},\textcolor{blue}{blue} and \textcolor{green}{green}.}
\label{tab:davis17_val}
\begin{tabular}{cccccccc}
\toprule
\multirow{2}{*}{\textbf{Method}} &
  \multirow{2}{*}{\textbf{BackBone}} &
  \multirow{2}{*}{\textbf{Training Dataset}} & 
  \multicolumn{5}{c}{\textbf{DAVIS17-val}} \\ \cmidrule(lr){4-8}
&  &  & \(\mathcal{J}_m\) & \(\mathcal{J}_r\) & \(\mathcal{F}_m\) & \(\mathcal{F}_r\) & \(\mathcal{J} \& \mathcal{F}_m\) \\ \midrule
MuG\cite{lu2020learning}&ResNet-18&OxUvA&52.6&57.4&56.1&58.1&54.3\\
MAST\cite{lai2020mast}&ResNet-18&YouTubeVOS&63.3&73.2&67.6&77.7&65.5\\
CRW\cite{jabri2020space}&ResNet-18&Kinetics&64.8&76.1&70.2&82.1&67.6\\
ContrastCorr\cite{wang2021contrastive}&ResNet-18&TrackingNet&60.5&-&65.5&-&63.0\\
VFS\cite{xu2021rethinking}&ResNet-18&Kinetics&64.0&-&69.4&-&66.7\\
JSTG\cite{zhao2021modelling}&ResNet-18&Kinetics&65.8&77.7&71.6&\textcolor{green}{\textbf{84.3}}&68.7\\
ODIN\cite{henaff2022object}&ResNet-50&ImageNet&54.3&-&53.9&-&54.1\\
SCC\cite{son2022contrastive}&ResNet-18&YouTubeVOS&67.4&78.8&\textcolor{green}{\textbf{73.6}}&\textcolor{blue}{\textbf{84.6}}&\textcolor{green}{\textbf{70.5}}\\
SFC\cite{hu2022semantic}&ResNet-18&YouTubeVOS&68.3&-&\textcolor{blue}{\textbf{74.0}}&-&\textcolor{blue}{\textbf{71.2}}\\
CLIP-S$^{4}$\cite{he2023clip}&ResNet-50&ImageNet&52.3&-&56.8&-&54.6\\
CrOC\cite{stegmuller2023croc}&ViT-S/16&COCO&56.5&-&60.2&-&58.4\\
VideoHiGraph\cite{qin2023exposing}&ResNet-18&Kinetics&\textcolor{green}{\textbf{67.9}}&\textcolor{green}{\textbf{80.7}}&73.1&83.8&\textcolor{green}{\textbf{70.5}}\\
RPT\cite{dave2024no}&ViT-B&Kinetics&60.5&-&63.6&-&62.1\\
RHMNet\cite{zhou2024reliability}&ResNet-50&ImageNet + COCO&66.2&-&68.6&-&67.4\\
SSPNet\cite{9953060}&ResNet-50&Kinetics&67.5&77.9&70.6&80.2&69.1\\
PIRVOS\cite{guo2025self}&ResNet-18&YouTubeVOS&\textcolor{blue}{\textbf{69.2}}&\textcolor{blue}{\textbf{82.5}}&71.6&84.0&70.4\\
\midrule
\rowcolor{gray!20} \textbf{SWT}&ResNet-18&COCO&\textcolor{red}{\textbf{70.5}}&\textcolor{red}{\textbf{83.8}}&\textcolor{red}{\textbf{76.1}}&\textcolor{red}{\textbf{86.8}}&\textcolor{red}{\textbf{73.3}}\\
\bottomrule
\end{tabular}
\end{table}

\section{Experiment}
\label{sec:experiments}
    All experiments use the implementation hyperparameters summarized in Table~\ref{tab:implementation_hyperparameters} unless otherwise specified.

    To validate our approach, we performed comparative experiments with advanced methods on multiple VOS benchmark datasets and additionally tested it on the body part propagation dataset VIP\cite{zhou2018adaptive}. We also experimentally present some of the test results, along with ablation studies for each module and further analyzes from multiple perspectives, fully demonstrating the superiority of our approach.
    
%===================================VOS results===================================
\subsection{Results for Video Object Segmentation}
\label{subsec:res_vos}
    \textbf{Dataset.} We evaluate our approach on five mainstream datasets: 1) DAVIS16\cite{perazzi2016benchmark}. A single-object VOS dataset consisting of 50 sequences, with 30 sequences for training and 20 for validation; 2) DAVIS17\cite{pont20172017}. A challenging multi-object VOS dataset comprising 90 sequences, with 60 for training and 30 for validation; 3) DAVIS17-test\cite{pont20172017}. The official test set from the VOS challenge, containing 30 multi-object VOS sequences where only the first-frame ground truth are provided; 4) YouTubeVOS18\cite{xu2018youtube}. Composed of 3,471 training videos and 474 validation videos, approximately half containing multiple objects. Objects are categorized into two classes: seen objects (present in both training and validation sets) and unseen objects (exclusively appearing in the validation set); 5) YouTubeVOS19\cite{yang2019video}. The 2019 version expands the dataset scale to more than 4,000 videos based on the 2018 version, it contains 507 validation videos.

    \textbf{Evaluation Metrics.} For thorough performance assessment, we utilize three standard evaluation metrics from the benchmark datasets: region similarity \(\mathcal{J}\), contour accuracy \(\mathcal{F}\), and their combined measure \(\mathcal{J} \& \mathcal{F}\):
    \begin{itemize}
        \item Region Similarity (\(\mathcal{J}\)): This metric computes the mean Intersection-over-Union (mIoU) between the predicted segmentation and the ground truth mask.
        \item Contour Accuracy (\(\mathcal{F}\)): This metric assesses boundary alignment quality by computing precision and recall rates of contour pixels between the predicted segmentation and ground truth mask.
        \item Overall Score (\(\mathcal{J} \& \mathcal{F}\)): The average of \(\mathcal{J}\) and \(\mathcal{F}\).
    \end{itemize}

    \begin{table}[t]
\centering
\setlength{\tabcolsep}{8.6pt}
\caption{Quantitative comparison to methods on DAVIS17-test. The three highest-scores are highlighted in colored text: \textcolor{red}{red},\textcolor{blue}{blue} and \textcolor{green}{green}.}
\label{tab:davis17_test}
\begin{tabular}{ccccc}
\toprule
\multirow{2}{*}{\textbf{Method}} & \multirow{2}{*}{\textbf{Supervised Mode}} & \multicolumn{3}{c}{\textbf{DAVIS17-test}} \\ \cmidrule{3-5} 
 &  & \(\mathcal{J}\) & \(\mathcal{F}\) & \(\mathcal{J} \& \mathcal{F}\) \\ \midrule
CRW\cite{jabri2020space} & Self-supervised & 52.3 & 59.6 & 55.9 \\
VFS\cite{xu2021rethinking} & Self-supervised & \textcolor{green}{\textbf{53.1}} & \textcolor{green}{\textbf{61.6}} & \textcolor{green}{\textbf{57.3}} \\
SCC\cite{son2022contrastive} & Self-supervised & \textcolor{blue}{\textbf{55.9}} & \textcolor{blue}{\textbf{64.0}} & \textcolor{blue}{\textbf{59.9}} \\ \midrule
\rowcolor{gray!20} \textbf{SWT(Ours)} & Self-supervised & \textcolor{red}{\textbf{57.4}} & \textcolor{red}{\textbf{66.1}} & \textcolor{red}{\textbf{61.8}} \\ \bottomrule
\end{tabular}
\end{table}
    \begin{table}[t]
\centering
\setlength{\tabcolsep}{7pt}
\caption{Quantitative comparison to methods on DAVIS16-val. The three highest-scores are highlighted in colored text: \textcolor{red}{red},\textcolor{blue}{blue} and \textcolor{green}{green}.}
\label{tab:davis16_val}
\begin{tabular}{ccccc}
\toprule
\multicolumn{1}{c}{\multirow{2}{*}{\textbf{Method}}} & \multirow{2}{*}{\textbf{Supervised Mode}} & \multicolumn{3}{c}{\textbf{DAVIS16-val}}                \\ \cmidrule{3-5} 
\multicolumn{1}{c}{}         &      & \(\mathcal{J}\)    & \(\mathcal{F}\)    & \multicolumn{1}{c}{\(\mathcal{J} \& \mathcal{F}\)} \\ \midrule
\textcolor{gray}{SiamMask\cite{wang2019fast}}& \textcolor{gray}{Semi-supervised} & \textcolor{gray}{71.7} & \textcolor{gray}{67.8} & \textcolor{gray}{70.0} \\
\textcolor{gray}{OSVOS\cite{caelles2017one}}& \textcolor{gray}{Semi-supervised} & \textcolor{gray}{79.8} & \textcolor{gray}{80.6}& \textcolor{gray}{80.2} \\

Colorization\cite{vondrick2018tracking} & Self-supervised & 38.9 & 30.0 & 34.9 \\
CorrFlow\cite{lai2019self} & Self-supervised & 47.1 & 49.9 & 48.0 \\
TimeCycle\cite{wang2019learning} & Self-supervised & 55.8 & 51.1 & 53.5 \\
MAST\cite{lai2020mast} & Self-supervised & 69.3 & 68.3 & 68.8 \\
MPFM\cite{li2022pixels} & Self-supervised & \textcolor{green}{\textbf{75.2}} & \textcolor{green}{\textbf{73.3}} & \textcolor{green}{\textbf{74.3}} \\
SPGO\cite{ponimatkin2023simple} & Self-supervised & \textcolor{red}{\textbf{80.2}} & \textcolor{blue}{\textbf{77.5}} & \textcolor{blue}{\textbf{78.8}} \\ \midrule
\rowcolor{gray!20} \textbf{SWT(Ours)} & Self-supervised & \textcolor{blue}{\textbf{79.6}} & \textcolor{red}{\textbf{81.3}} & \textcolor{red}{\textbf{80.5}} \\

%\textcolor{gray}{STM\cite{oh2019video}}& \textcolor{gray}{Semi-supervised} & \textcolor{gray}{88.7} & \textcolor{gray}{89.9} & \textcolor{gray}{89.3} \\ 
\bottomrule
\end{tabular}
\end{table}   
    
    \textbf{Evaluation on DAVIS17.} As shown in Table \ref{tab:davis17_val}, we compare our method with advanced  self-supervised VOS approaches on the DAVIS17-val dataset. Our SWT achieves \textbf{73.3\%} \(\mathcal{J} \& \mathcal{F}\), outperforming all other methods across all metrics. Notably, compared to recent methods\cite{guo2025self}, we improve \(\mathcal{F}\) by 4.5\% and \(\mathcal{J} \& \mathcal{F}\) by 2.9\%. Even when compared to methods with stronger backbones (e.g. ODIN\cite{henaff2022object}, CLIP-S$^{4}$\cite{he2023clip}, CrOC\cite{stegmuller2023croc}, RPT\cite{dave2024no}), our approach surpasses them using only ResNet-18\cite{he2016deep} as the backbone. To further ensure fairness in evaluation, we also conduct comparisons on the DAVIS17-test dataset. As shown in Table \ref{tab:davis17_test}, our method maintains its leading performance even on this more challenging dataset.

    \textbf{Evaluation on DAVIS16.} To complete the evaluation criteria, we also conduct comparisons on the DAVIS16-val dataset. As shown in Table \ref{tab:davis16_val}, our method performs excellently on the single-object dataset. Compared with SPGO\cite{ponimatkin2023simple} that uses additional optical flow as auxiliary information, our method improves \(\mathcal{J} \& \mathcal{F}\) by 1.7\%. Moreover, our method even outperforms some semi-supervised VOS methods (e.g. OSVOS\cite{caelles2017one}, SiamMask\cite{wang2019fast}).

    \textbf{Evaluation on YouTubeVOS.} We also conduct comparisons on the YouTubeVOS dataset. The results on the YouTubeVOS18-val dataset are shown in Table \ref{tab:ytb_18}. Compared to LIIR\cite{li2022locality}, we achieve a 2.7\% improvement in mean score. Furthermore, our method surpasses some well-known semi-supervised VOS methods (e.g. OSVOS\cite{caelles2017one}, PreMVOS\cite{luiten2018premvos}, TVOS\cite{zhang2020transductive}). High scores on both seen and unseen categories demonstrate the robustness and adaptability of our approach. And our experimental evaluation on the YouTubeVOS19-val (Table \ref{tab:ytb_19}) demonstrates advanced performance, with the quantitative results confirming the effectiveness of our approach.

    \begin{table}[t]
\centering
\setlength{\tabcolsep}{2.3pt}
\caption{Quantitative comparison to methods on YoutubeVOS18-val set. The three highest-scores are highlighted in colored text: \textcolor{red}{red},\textcolor{blue}{blue} and \textcolor{green}{green}.}
\label{tab:ytb_18}
\begin{tabular}{ccccccc}
\toprule
\multirow{2}{*}{\textbf{Method}} & \multirow{2}{*}{\textbf{Supervised Mode}} & \multicolumn{1}{l}{\multirow{2}{*}{\textbf{Mean}}} & \multicolumn{2}{c}{\textbf{Seen}} & \multicolumn{2}{l}{\textbf{Unseen}} \\ \cmidrule{4-7} 
 &  & \multicolumn{1}{l}{} & \(\mathcal{J}\) & \(\mathcal{F}\) & \multicolumn{1}{l}{\(\mathcal{J}\)} & \multicolumn{1}{l}{\(\mathcal{F}\)} \\ \midrule
\textcolor{gray}{OSVOS\cite{caelles2017one}} & \textcolor{gray}{Semi-supervised} & \textcolor{gray}{58.8} & \textcolor{gray}{59.8} & \textcolor{gray}{60.5} & \textcolor{gray}{54.2} & \textcolor{gray}{60.7} \\
\textcolor{gray}{PreMVOS\cite{luiten2018premvos}} & \textcolor{gray}{Semi-supervised} & \textcolor{gray}{66.9} & \textcolor{gray}{71.4} & \textcolor{gray}{75.9} & \textcolor{gray}{56.5} & \textcolor{gray}{63.7} \\ 
\textcolor{gray}{TVOS\cite{zhang2020transductive}} & \textcolor{gray}{Semi-supervised} & \textcolor{gray}{67.8} & \textcolor{gray}{67.1} & \textcolor{gray}{69.4} & \textcolor{gray}{63.0} & \textcolor{gray}{71.6} \\ 
 
Colorization\cite{vondrick2018tracking} & Self-supervised & 38.9 & 43.1 & 38.6 & 36.6 & 37.4 \\
CorrFlow\cite{lai2019self} & Self-supervised & 46.6 & 50.6 & 46.6 & 43.8 & 45.7 \\
RHMNet\cite{zhou2024reliability} & Scribble-supervised & 59.6 & 64.9 & 65.8 & 51.3 & 56.6 \\
MAST\cite{lai2020mast} & Self-supervised & 64.2 & 64.9 & 64.9 & 60.3 & 67.7 \\
SSPNet\cite{9953060} & Self-supervised & 65.9 & 65.7 & 67.0 & 61.9 & 69.1 \\
CLTC\cite{jeon2021mining} & Self-supervised & 67.3 & 66.2 & \textcolor{green}{\textbf{67.9}} & 63.2 & 71.7 \\
PIRVOS\cite{guo2025self} & Self-supervised & \textcolor{green}{\textbf{68.5}} & \textcolor{green}{\textbf{67.7}} & \textcolor{blue}{\textbf{69.2}} & \textcolor{green}{\textbf{64.4}} & \textcolor{green}{\textbf{72.5}} \\
LIIR\cite{li2022locality} & Self-supervised & \textcolor{blue}{\textbf{69.3}} & \textcolor{blue}{\textbf{67.9}} & \textcolor{red}{\textbf{69.7}} & \textcolor{blue}{\textbf{65.7}} & \textcolor{blue}{\textbf{ 73.8 }}\\ \midrule
\rowcolor{gray!20} \textbf{SWT(Ours)} & Self-supervised & \textcolor{red}{\textbf{72.0}} & \textcolor{red}{\textbf{70.3}} & 67.0 & \textcolor{red}{\textbf{72.8}} & \textcolor{red}{\textbf{77.7}} \\ \bottomrule
\end{tabular}
\end{table}

    \begin{table}[t]
\centering
\setlength{\tabcolsep}{2.8pt}
\caption{Quantitative comparison to methods on YoutubeVOS19-val set. The three highest-scores are highlighted in colored text: \textcolor{red}{red},\textcolor{blue}{blue} and \textcolor{green}{green}.}
\label{tab:ytb_19}

\begin{tabular}{ccccccc}
\toprule
\multirow{2}{*}{\textbf{Method}} & \multirow{2}{*}{\textbf{Supervised Mode}} & \multirow{2}{*}{\textbf{Mean}} & \multicolumn{2}{c}{\textbf{Seen}} & \multicolumn{2}{c}{\textbf{Unseen}} \\ \cmidrule{4-7} 
 &  &  & \(\mathcal{J}\) & \(\mathcal{F}\) & \(\mathcal{J}\) & \(\mathcal{F}\) \\ \midrule
Colorization\cite{vondrick2018tracking} & Self-supervised & 39.0 & 43.3 & 38.2 & 36.6 & 37.5 \\
CorrFlow\cite{lai2019self} & Self-supervised & 47.0 & 51.2 & 46.6 & 44.5 & 45.9 \\
RHMNet\cite{zhou2024reliability} & Scribble-supervised & \textcolor{green}{\textbf{59.3}} & \textcolor{green}{\textbf{63.9}} & \textcolor{green}{\textbf{64.8}} & \textcolor{green}{\textbf{51.7}} & \textcolor{green}{\textbf{56.7}} \\
MAST\cite{lai2020mast} & Self-supervised & \textcolor{blue}{\textbf{64.9}} & \textcolor{blue}{\textbf{64.3}} & \textcolor{blue}{\textbf{65.3}} & \textcolor{blue}{\textbf{61.5}} & \textcolor{blue}{\textbf{68.4}} \\ \midrule
\rowcolor{gray!20} \textbf{SWT(Ours)} & Self-supervised & \textcolor{red}{\textbf{71.7}} & \textcolor{red}{\textbf{69.5}} & \textcolor{red}{\textbf{67.7}} & \textcolor{red}{\textbf{71.6}} & \textcolor{red}{\textbf{77.8}} \\ \bottomrule
\end{tabular}
\end{table}

%===================================BPP results===================================
\subsection{Results for Body Part Propagation}
\label{subsec:res_bpp}

    Part-focused feature learning has been explored for anthropometric measurement~\cite{chen2025focused}; here, we evaluate fine-grained human-body correspondence through body part propagation.

    \textbf{Dataset.} We evaluate SWT on the Video Instance Parsing (VIP) dataset\cite{zhou2018adaptive} for body part propagation. The val-set split of this benchmark contains 50 videos focusing on propagating 19 body parts such as arms and legs. Consequently, this task demands higher matching precision compared to video object segmentation. We adopt the same settings as CRW\cite{jabri2020space} and resize video frames to \(560 \times 560\).

    \textbf{Evaluation metrics.} To evaluate our model's performance on the propagation task, we use the standard metric provided by\cite{zhou2018adaptive}, namely mean Intersection-over-Union (mIoU).

    \textbf{Evaluation on VIP.} Table \ref{tab:vip} compares the performance of our model with existing self-supervised methods. In terms of mIoU, our SWT outperforms LIIR\cite{li2022locality}  and SCC\cite{son2022contrastive} by 1.9\% and 2.3\% respectively. Moreover, SWT shows a significant 5.2\% improvement over the fully-supervised model ATEN\cite{zhou2018adaptive}, which was specifically designed for the VIP dataset.

    \begin{table}[t]
\centering
\setlength{\tabcolsep}{16.1pt}
\caption{Quantitative results for body part propagation on VIP val set. The three highest-scores are highlighted in colored text: \textcolor{red}{red},\textcolor{blue}{blue} and \textcolor{green}{green}.}
\label{tab:vip}
\begin{tabular}{ccc}
\toprule
\multirow{2}{*}{\textbf{Method}} & \multirow{2}{*}{\textbf{Supervised Mode}} & \multicolumn{1}{c}{\textbf{VIP-val}}\\ \cmidrule{3-3} 
 &  & mIoU \\ \midrule
\textcolor{gray}{ResNet\cite{he2016deep}} & \textcolor{gray}{Fully-supervised} & \textcolor{gray}{31.9} \\
\textcolor{gray}{ATEN\cite{zhou2018adaptive}} & \textcolor{gray}{Fully-supervised} & \textcolor{gray}{37.9} \\

TimeCycle\cite{wang2019learning} & Self-supervised & 28.9 \\
UVC\cite{li2019joint} & Self-supervised & 34.1 \\
CLTC\cite{jeon2021mining} & Self-supervised & 37.8 \\
CRW\cite{jabri2020space} & Self-supervised & 38.6 \\
VFS\cite{xu2021rethinking} & Self-supervised & 39.9 \\
SCC\cite{son2022contrastive} & Self-supervised & \textcolor{green}{\textbf{40.8}} \\
LIIR\cite{li2022locality} & Self-supervised & \textcolor{blue}{\textbf{41.2}} \\ \midrule
\rowcolor{gray!20} \textbf{SWT(ours)} & Self-supervised & \textcolor{red}{\textbf{43.1}} \\ \bottomrule
\end{tabular}
\end{table}

    \begin{table}[t]
\centering
\setlength{\tabcolsep}{5.6pt}
\caption{Ablation studies on individual modules. The three highest-scores are highlighted in colored text: \textcolor{red}{red},\textcolor{blue}{blue} and \textcolor{green}{green}.}
\label{tab:ablation_main}
\begin{tabular}{cccccc}
\toprule
\multirow{2}{*}{\textbf{Module Variants}} & \multirow{2}{*}{\textbf{S}} & \multirow{2}{*}{\textbf{W}} & \multirow{2}{*}{\textbf{T}} & \textbf{DAVIS17-val} & \textbf{DAVIS16-val} \\ \cmidrule{5-6} 
          &   &   &   & \(\mathcal{J} \& \mathcal{F}\) & \(\mathcal{J} \& \mathcal{F}\) \\ \midrule
Baseline  & \textcolor{red}{\text{\ding{55}}} & \textcolor{red}{\text{\ding{55}}} & \textcolor{red}{\text{\ding{55}}} & 72.0 & 78.9          \\
S         & \textcolor{green}{\checkmark } & \textcolor{red}{\text{\ding{55}}} & \textcolor{red}{\text{\ding{55}}} & 72.3 & 79.3\\
W         & \textcolor{red}{\text{\ding{55}}} & \textcolor{green}{\checkmark} & \textcolor{red}{\text{\ding{55}}} & 72.6 & 79.6\\
T         & \textcolor{red}{\text{\ding{55}}} & \textcolor{red}{\text{\ding{55}}} & \textcolor{green}{\checkmark } & \textcolor{blue}{\textbf{73.1}} & \textcolor{green}{\textbf{80.2}} \\
S + W     & \textcolor{green}{\checkmark } & \textcolor{green}{\checkmark } & \textcolor{red}{\text{\ding{55}}} & \textcolor{green}{\textbf{73.0}} & \textcolor{blue}{\textbf{80.3}} \\
S + T     & \textcolor{green}{\checkmark } & \textcolor{red}{\text{\ding{55}}} & \textcolor{green}{\checkmark } & \textcolor{blue}{\textbf{73.1}} & 80.0\\
W + T     & \textcolor{red}{\text{\ding{55}}} & \textcolor{green}{\checkmark } & \textcolor{green}{\checkmark } & \textcolor{green}{\textbf{73.0}} & 80.1 \\ \midrule
\rowcolor{gray!20} S + W + T & \textcolor{green}{\checkmark } & \textcolor{green}{\checkmark } & \textcolor{green}{\checkmark } & \textcolor{red}{\textbf{73.3}} & \textcolor{red}{\textbf{80.5}} \\ \bottomrule
\end{tabular}
\end{table}

    \begin{table}[t]
\centering
\setlength{\tabcolsep}{8.6pt}
\caption{Ablation study on augmentation strategies and marginal distributions. The highest score is shown in bold.}
\label{tab:ablation_crop}
\begin{tabular}{ccc}
\toprule
\multicolumn{1}{l}{\multirow{2}{*}{\textbf{Augmentation Strategy}}} & \textbf{DAVIS17-val} & \textbf{DAVIS16-val} \\ \cmidrule{2-3} 
 & \(\mathcal{J} \& \mathcal{F}\) & \(\mathcal{J} \& \mathcal{F}\) \\ \midrule
Detection Cropping & 71.9 & 78.8 \\
Enhanced Detection Cropping & 72.8 & 79.9 \\ 
\rowcolor{gray!20} Random Cropping & \textbf{73.3} & \textbf{80.5} \\ \midrule
\multicolumn{3}{l}{\textbf{Marginal Distributions}} \\ \midrule
Uniform & 73.2 & 80.4 \\
\rowcolor{gray!20} Cross Correlation & \textbf{73.3} & \textbf{80.5}\\ 
\bottomrule
\end{tabular}
\end{table}

    \begin{table*}[t]
\centering
\normalsize
\setlength{\tabcolsep}{5pt}
\caption{Sensitivity ablations for wavelet types, OT cost functions, sliding-window parameters, and temporal-loss weight \(\alpha\).}
\label{tab:ablation_sensitivity}
\begin{minipage}[t]{0.48\textwidth}
\vspace{0pt}
\centering
\textbf{(a) Wavelet Type}\\[2pt]
\begin{tabular}{lcc}
\toprule
\textbf{Type} & \textbf{DAVIS17-val} & \textbf{DAVIS16-val} \\
\midrule
Haar & \textbf{73.3} & \textbf{80.5} \\
Symlet-2 & 72.7 & 79.9 \\
Daubechies-2 & 71.7 & 77.4 \\
\bottomrule
\end{tabular}
\end{minipage}
\hfill
\begin{minipage}[t]{0.48\textwidth}
\vspace{0pt}
\centering
\textbf{(b) OT Cost}\\[2pt]
\begin{tabular}{lcc}
\toprule
\textbf{Cost} & \textbf{DAVIS17-val} & \textbf{DAVIS16-val} \\
\midrule
Euclidean & \textbf{73.3} & \textbf{80.5} \\
Cosine & 73.1 & 80.2 \\
\bottomrule
\end{tabular}
\end{minipage}
\par\medskip
\begin{minipage}[t]{0.48\textwidth}
\vspace{0pt}
\centering
\textbf{(c) Sliding Window}\\[2pt]
\begin{tabular}{lcc}
\toprule
\textbf{\((c,N)\)} & \textbf{DAVIS17-val} & \textbf{DAVIS16-val} \\
\midrule
\rowcolor{gray!20}
\multicolumn{3}{c}{\textit{Fixed capacity: \(cN=64\)}} \\
\midrule
\((1,64)\) & 72.0 & 78.4 \\
\((2,32)\) & 72.1 & 78.8 \\
\((4,16)\) & \textbf{72.3} & \textbf{79.2} \\
\((8,8)\) & 72.1 & 78.7 \\
\midrule
\rowcolor{gray!20}
\multicolumn{3}{c}{\textit{Fixed buffer length: \(c=4\)}} \\
\midrule
\((4,4)\) & 71.1 & 77.9 \\
\((4,8)\) & 71.6 & 78.5 \\
\((4,16)\) & \textbf{72.3} & \textbf{79.2} \\
\bottomrule
\end{tabular}
\end{minipage}
\hfill
\begin{minipage}[t]{0.48\textwidth}
\vspace{0pt}
\centering
\textbf{(d) Temporal-Loss Weight}\\[2pt]
\begin{tabular}{lcc}
\toprule
\textbf{\(\alpha\)} & \textbf{DAVIS17-val} & \textbf{DAVIS16-val} \\
\midrule
0 & 71.9 & 77.6 \\
0.25 & 72.4 & 78.7 \\
0.5 & 72.2 & 78.3 \\
1 & \textbf{73.3} & \textbf{80.5} \\
\bottomrule
\end{tabular}
\end{minipage}
\vspace{2pt}
\end{table*}

    \begin{table}[t]
\centering
\setlength{\tabcolsep}{4pt}
\caption{Efficiency comparison between the baseline and SWT.}
\label{tab:efficiency_analysis}
\begin{tabular}{lcccc}
\toprule
\textbf{Method} & \textbf{Params} & \textbf{FLOPs} & \textbf{GPU memory} & \textbf{FPS} \\
\midrule
Baseline & 0.753M & 1.865G & 16.0GB & 48 \\
SWT & 1.160M & 2.064G & 21.4GB & 43 \\
\bottomrule
\end{tabular}
\end{table}

%===================================ablation studies===================================
\subsection{Ablation Studies}
\label{subsec:ablation}
    To comprehensively evaluate our method, we conduct systematic ablation studies. Specifically, we design eight sets of experiments based on the three proposed modules (\textbf{S}liding Window, \textbf{W}avelet Transform, and Optimal \textbf{T}ransport) and test them on DAVIS17-val\cite{pont20172017} and DAVIS16-val\cite{perazzi2016benchmark} to assess each module's contribution. Additionally, we perform ablation studies on static image augmentation and marginal distribution initialization to further analyze our proposed approach.

    \textbf{Sensitivity Analysis.} We further examine the sensitivity of SWT to wavelet types, OT cost functions, sliding-window parameters, and the temporal-loss weight \(\alpha\) in Table \ref{tab:ablation_sensitivity}. For the wavelet encoder, Haar achieves the best results on both DAVIS17-val and DAVIS16-val, outperforming Symlet-2 and Daubechies-2. For optimal transport, the Euclidean cost obtains slightly better performance than the cosine cost, so we keep it as the default transport cost. For the sample buffer, the fixed-capacity experiment uses \(cN=64\) and varies \((c,N)\) among \((1,64)\), \((2,32)\), \((4,16)\), and \((8,8)\); \((4,16)\) performs best in this comparison. To isolate the refresh size \(N\), we additionally fix \(c=4\) and evaluate \(N=4,8,16\). The corresponding results increase from 71.1\%/77.9\% to 71.6\%/78.5\% and 72.3\%/79.2\% on DAVIS17-val/DAVIS16-val, supporting \(N=16\) as the default among the tested values. For the hybrid loss, removing the temporal term with \(\alpha=0\) yields 71.9\% and 77.6\% on DAVIS17-val and DAVIS16-val, respectively. The intermediate settings \(\alpha=0.25\) and \(\alpha=0.5\) improve upon \(\alpha=0\), although the trend is not strictly monotonic. Equal weighting with \(\alpha=1\) achieves the best results of 73.3\% and 80.5\%, indicating that the temporal term contributes to the final performance and supporting \(\alpha=1\) as the default setting.

    \textbf{Sliding-Window Sample Buffer.} We evaluate the rolling buffer with \(c=4\) in Table \ref{tab:ablation_main} (row 2). It improves \(\mathcal{J} \& \mathcal{F}\) by 0.3\% on DAVIS17-val and 0.4\% on DAVIS16-val. These modest gains indicate that overlapping sample reuse can benefit optimization without introducing additional training images. Accordingly, the buffer serves as a supporting sample-reuse training strategy rather than the primary modeling innovation; the main modeling components are the wavelet transform and optimal transport, which address long-range context aggregation and globally constrained correspondence, respectively.
    
    \begin{figure}[t]
      \centering
        \includegraphics[width=0.95\linewidth]{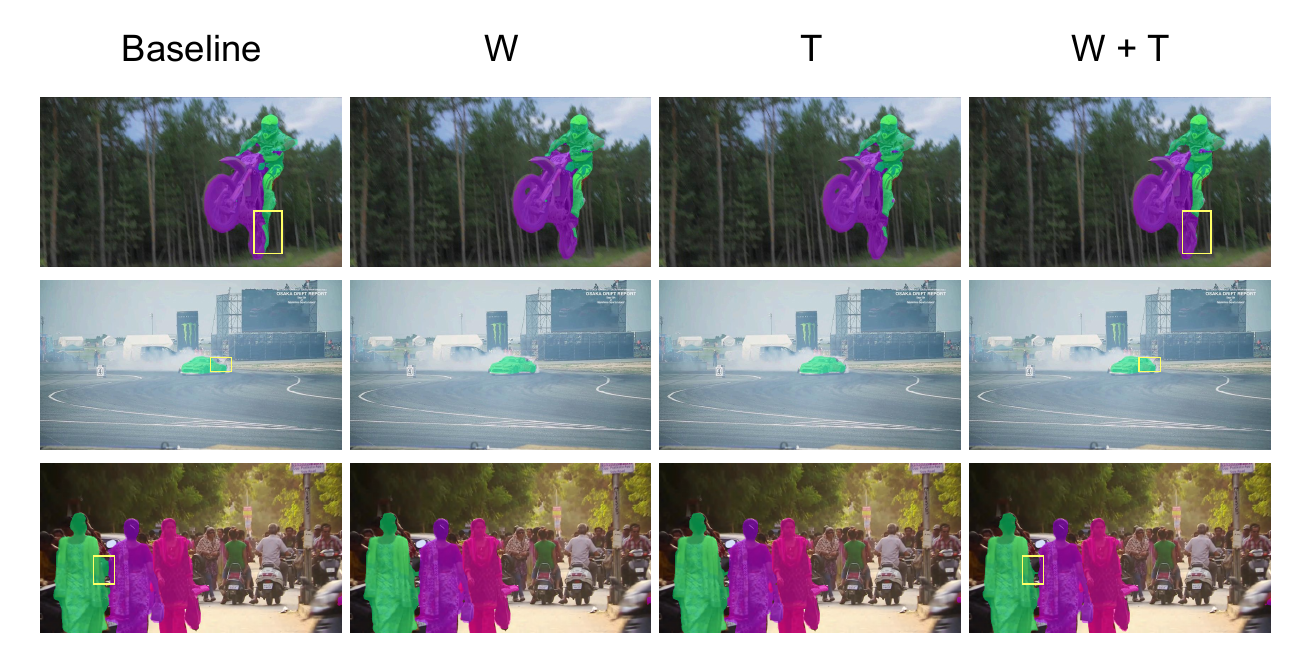}
        \caption{
            Ablation results of the wavelet transform and optimal transport modules. The four columns from left to right correspond to: baseline, adding only the wavelet transform module, adding only the optimal transport module, and incorporating both modules, respectively. The yellow boxes circle the area of accuracy improvement.
        }
        \label{fig:ablation_davis}
    \end{figure}

    \begin{figure}[!t]
    \centering
    \begin{subfigure}{0.65\textwidth}
        \centering
        \includegraphics[width=\linewidth]{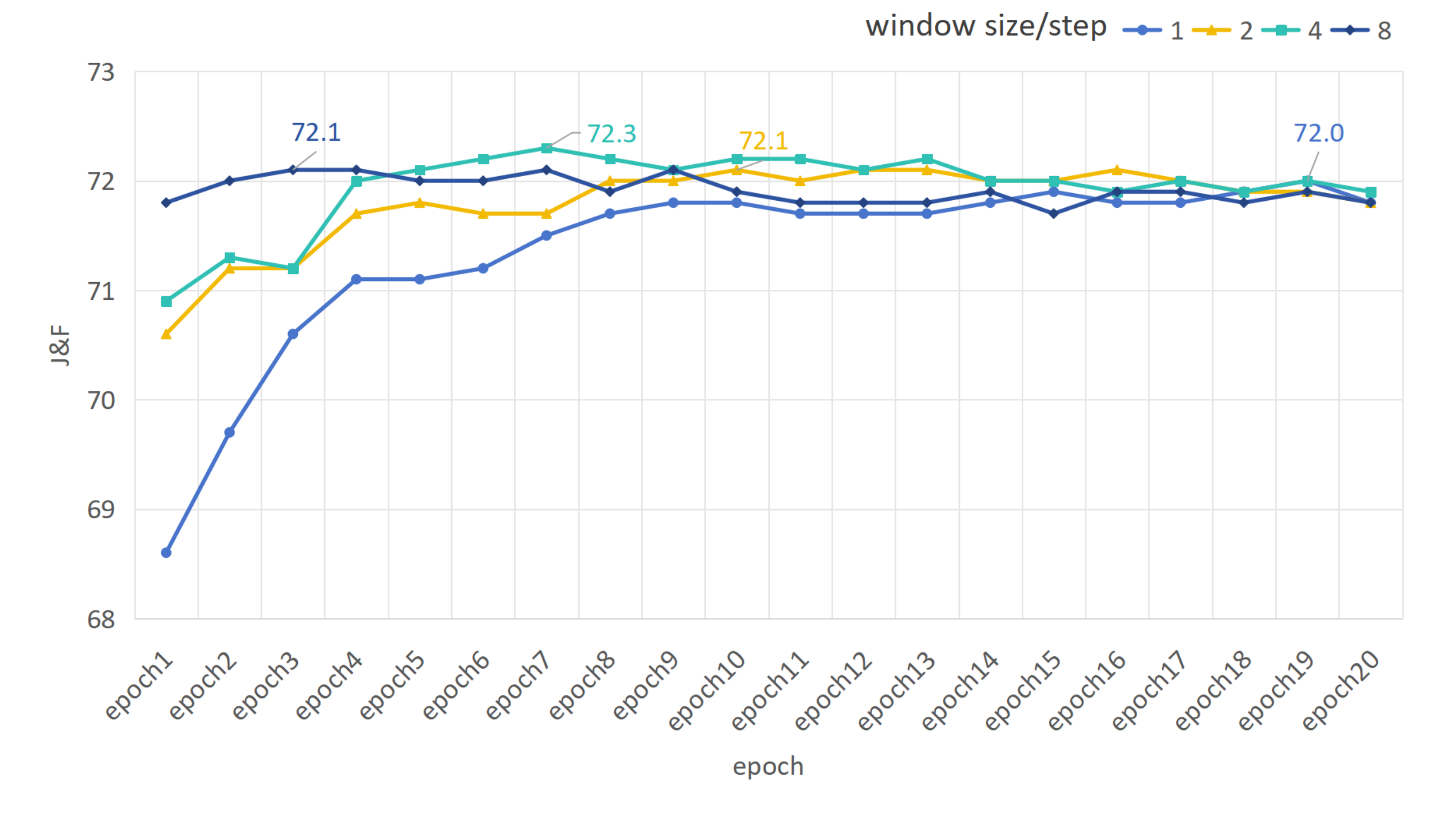}
        \caption{
            Convergence for \((c,N)=(1,64),(2,32),(4,16),(8,8)\) at a fixed buffer capacity \(cN=64\). The horizontal axis represents training epochs, while the vertical axis shows \(\mathcal{J} \& \mathcal{F}\).
        }
        \label{fig:ana_s}
    \end{subfigure}
    \par\medskip
    \begin{subfigure}{0.65\textwidth}
         \centering
        \includegraphics[width=\linewidth]{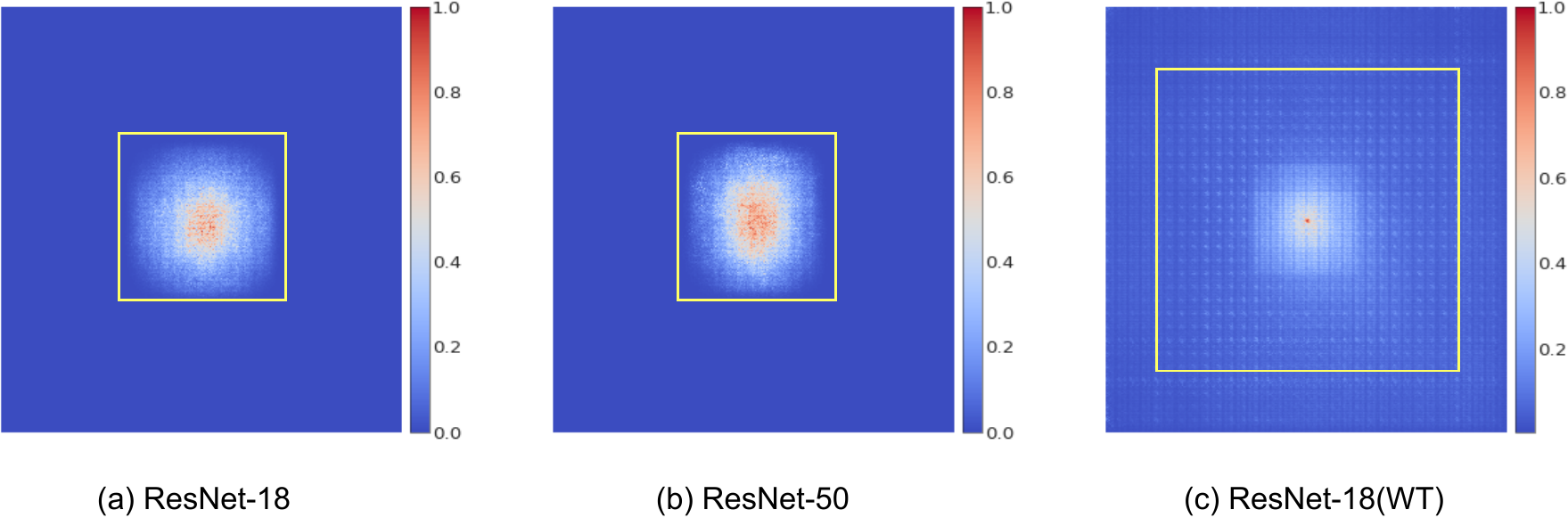}
        \caption{
            Visualization of effective receptive fields for different backbones. From (a) to (c) are the effective receptive fields of ResNet-18, ResNet-50, and ResNet-18 (WT) respectively. The yellow boxes are the effective sensory field range.
        }
        \label{fig:ana_w}
    \end{subfigure}
    \caption{Further analysis of convergence under different sliding-window buffer settings and effective receptive fields under different backbone configurations.}
    \end{figure}

    % \begin{figure}[t]
    %     \centering
    %     \includegraphics[width=\linewidth]{pics/analysis_S_v2.pdf}
    %     \caption{
    %         The convergence performance of models under different sliding window sizes/step. The horizontal axis represents training epochs, while the vertical axis shows \(\mathcal{J} \& \mathcal{F}\) metrics.
    %     }
    %     \label{fig:ana_s}
    % \end{figure}
    
    % \begin{figure}[t]
    %     \centering
    %     \includegraphics[width=\linewidth]{pics/analysis_W_v2.pdf}
    %     \caption{
    %         Visualization of effective receptive fields for different backbones. From (a) to (c) are the effective receptive fields of ResNet-18, ResNet-50, and ResNet-18 (WT) respectively. The yellow boxes are the effective sensory field range.
    %     }
    %     \label{fig:ana_w}
    % \end{figure}

    \textbf{Wavelet Transform.} As shown in row 3 of Table \ref{tab:ablation_main}, we conduct independent validation experiments on the wavelet transform encoder. The results demonstrate 0.6\% and 0.7\% \(\mathcal{J} \& \mathcal{F}\) improvements on DAVIS17-val and DAVIS16-val datasets respectively. This improvement verifies the effectiveness of wavelet-domain convolution operations: by expanding the receptive field for long-range context modeling, robust feature extraction can be achieved even with simple convolutional architectures. As illustrated in column 2 of Figure \ref{fig:ablation_davis}, compared to the baseline, the wavelet transform significantly enhances output mask refinement.

    \textbf{Optimal Transport.} As shown in row 4 of Table \ref{tab:ablation_main}, we conduct independent validation experiments for the model incorporating the optimal transport module. Quantitative analysis reveals this module achieves 1.1\% and 1.3\% \(\mathcal{J} \& \mathcal{F}\) improvements on DAVIS17-val and DAVIS16-val datasets respectively. This significant performance gain validates the applicability of optimal transport theory for self-supervised VOS tasks: its superior global modeling capability effectively handles challenges like rapid object motion and deformation, perfectly aligning with core VOS requirements. Qualitative results (Figure \ref{fig:ablation_davis}, column 3) further demonstrate the module's pronounced improvement on output mask boundary precision.

    \textbf{Component Interaction.} The gains of the components in Table \ref{tab:ablation_main} are not strictly additive. The T variant obtains 73.1\% and 80.2\% on DAVIS17-val and DAVIS16-val, respectively, compared with 73.0\% and 80.1\% for W+T and 73.1\% and 80.0\% for S+T. Optimal transport directly constrains global correspondence over the current feature distributions, whereas W changes the receptive field and frequency characteristics of the features and S changes sample reuse across optimization steps. Under the same fixed training settings, these changes also alter the feature and gradient statistics on which the OT objective operates, so the pairwise variants can converge to slightly different solutions. The resulting differences of 0.0--0.2\% are small and are not interpreted as evidence of systematic conflict between the components. Under the reported configuration, the complete S+W+T model reaches 73.3\% and 80.5\%, outperforming T alone by 0.2\% and 0.3\% and achieving the best overall results.

    \textbf{Design Alternatives.} We further discuss simpler alternatives that may appear to provide similar effects. Larger or more object-centric crops can increase foreground coverage, but our detection-cropping augmentation study shows that detection-guided crops perform worse than random crops, likely because over-tight boxes reduce background context and generate unrealistic motion patterns. Dilated convolutions can enlarge the receptive field, but they do not provide the frequency decomposition and low-frequency contextual enhancement offered by wavelet-domain convolution. Standard attention or non-local matching can aggregate pairwise similarities, yet it lacks the marginal distribution constraints of optimal transport and therefore does not explicitly enforce globally balanced soft assignment. The ablation results also show that the OT module provides the largest single-module gain among the three components, while the full SWT model achieves the best overall performance. We therefore retain OT and wavelet-domain modeling as the main global correspondence components rather than replacing them with these simpler alternatives.

    \textbf{Augmentation Strategy.} Intuitively, we hypothesize that when cropping two frames from static images as model input, we should prioritize regions containing salient objects to help the model focus on learning object features in videos. We initially employ YOLO\cite{wang2023yolov7} to detect the highest-confidence object bounding box in static images, using this box as the first frame, then slightly scaling and shifting the box to crop the second frame. However, this detection cropping strategy performs worse than the baseline. We attribute this to: 1) over-tight object boxes causing the model to focus solely on foreground while ignoring background information, and 2) unrealistic motion patterns from simple box shifting.We then enhance this approach by first randomly enlarging detected object boxes before cropping (to include more background), then moving boxes within a circular area constrained by average object motion pixels from real videos (to simulate realistic motion), combined with minor scaling. However, this enhanced detection cropping still underperforms. Our final method therefore retains random cropping, with comparative results shown in Table \ref{tab:ablation_crop}.

    \textbf{Marginal Distribution.} To further analyze the impact of marginal distribution initialization strategies, we conduct systematic comparative experiments between the two initialization methods proposed in Eq. \ref{eq:cc} and Eq. \ref{eq:uniform} (cross-correlation initialization vs. uniform distribution initialization). As shown in Table \ref{tab:ablation_crop}, quantitative results demonstrate that the cross-correlation-based initialization slightly outperforms uniform initialization in performance metrics. This discrepancy validates the effectiveness of leveraging feature correlations for distribution initialization within the optimal transport.

    \begin{figure}[t]
    \centering
    \begin{subfigure}{0.49\textwidth}
         \centering
        \includegraphics[width=0.95\linewidth]{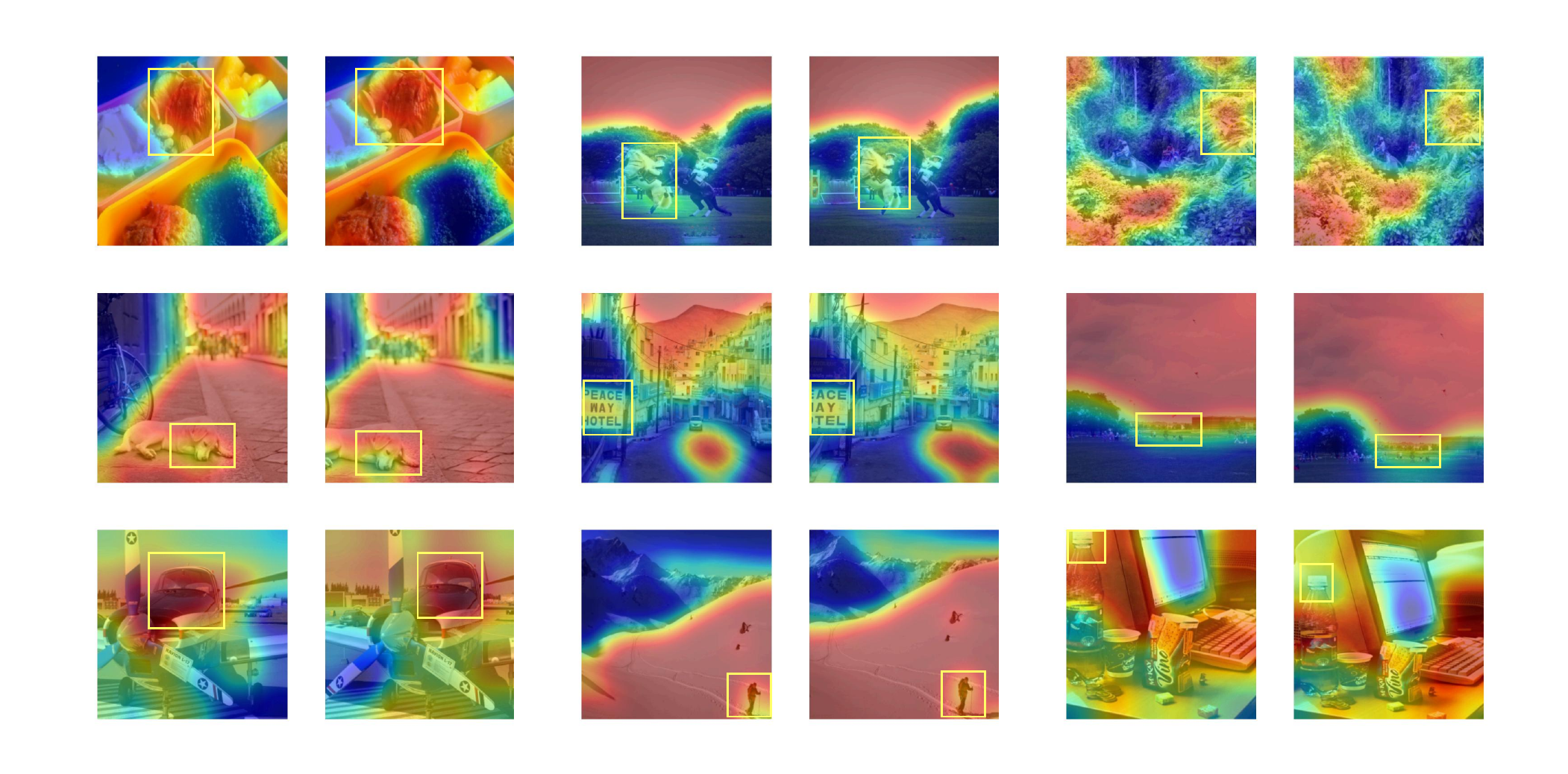}
        \caption{
            Visualization of cross-correlation results. The heatmap of cross-correlation results is shown, where darker colors indicate higher attention weights. The yellow boxes circled are cross-frame associations.
        }
        \label{fig:ana_T_hm}
    \end{subfigure}
    \hfill
    \begin{subfigure}{0.49\textwidth}
        \centering
        \includegraphics[width=0.76\linewidth]{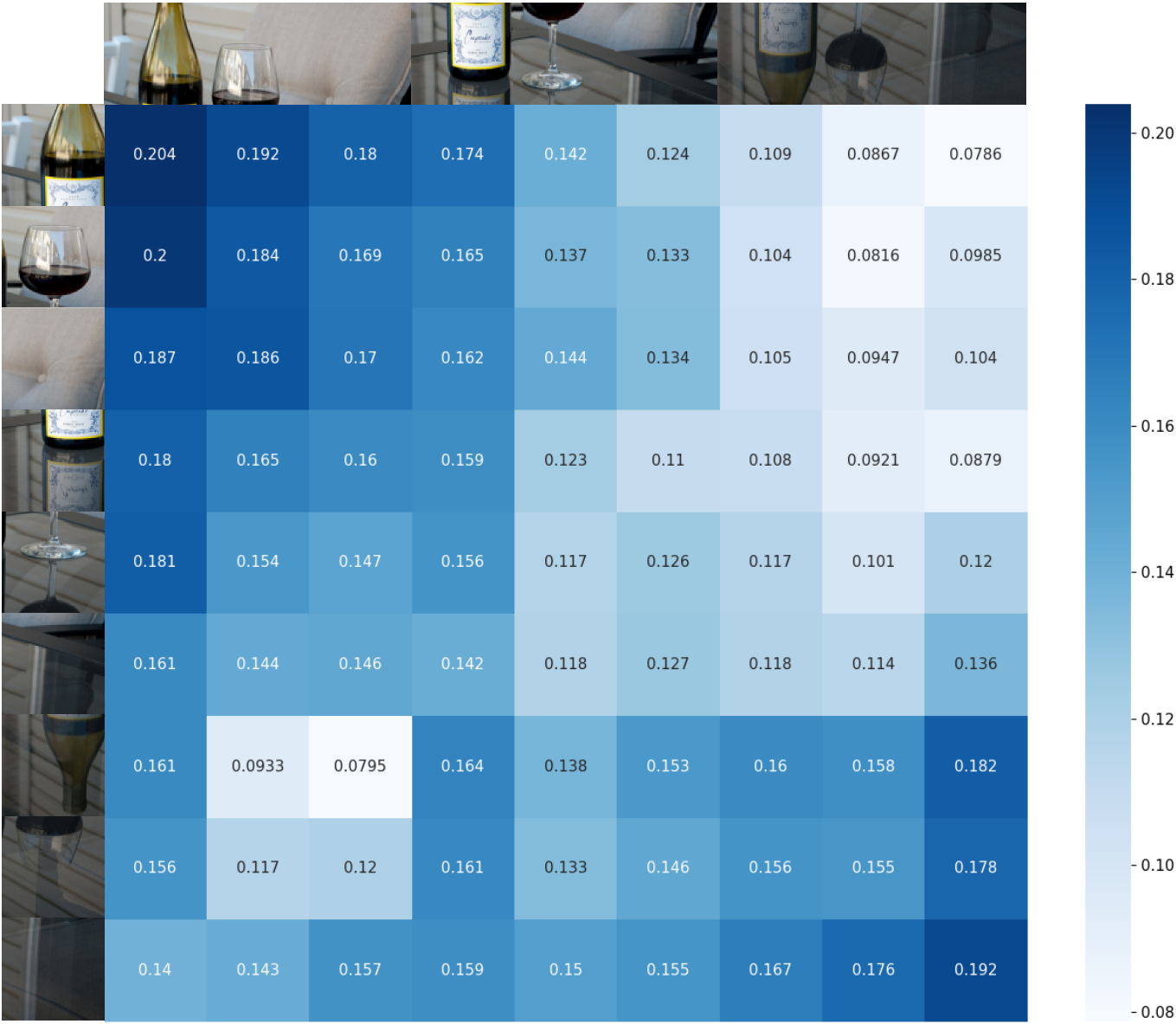}
        \caption{
            The similarity between two views captured by optimal transport. The first column on the left corresponds to patches generated from the first view, while the top row represents patches generated from the second view. The element at row \(i\) and column \(j\) in the matrix indicates the similarity between the \(i^{th}\) patch in view 1 and the \(j^{th}\) patch in view 2.
        }
        \label{fig:ana_T_cm}
    \end{subfigure}
    \caption{This is a further experimental analysis of the effectiveness of Optimal Transport.}
    \end{figure}
        
    %  \begin{figure}[t]
    %     \centering
    %     \includegraphics[width=\linewidth]{pics/analysis_T_hm_v3.pdf}
    %     \caption{
    %         Visualization of cross-correlation results. The heatmap of cross-correlation results is shown, where darker colors indicate higher attention weights. The yellow boxes circled are cross-frame associations.
    %     }
    %     \label{fig:ana_T_hm}
    % \end{figure}
    
    % \begin{figure}[t]
    %     \centering
    %     \includegraphics[width=0.75\linewidth]{pics/analysis_T_cm.pdf}
    %     \caption{
    %         The similarity between two views captured by optimal transport. The first column on the left corresponds to patches generated from the first view, while the top row represents patches generated from the second view. The element at row \(i\) and column \(j\) in the matrix indicates the similarity between the \(i^{th}\) patch in view 1 and the \(j^{th}\) patch in view 2.
    %     }
    %     \label{fig:ana_T_cm}
    % \end{figure}

    \textbf{Efficiency Analysis.} To evaluate the practicality and scalability of SWT, we report the computational cost, GPU memory usage, and inference speed in Table \ref{tab:efficiency_analysis}. Compared with the baseline, SWT increases the number of parameters from 0.753M to 1.160M and the FLOPs from 1.865G to 2.064G. The GPU memory usage increases from 16.0GB to 21.4GB, while the inference speed decreases moderately from 48 FPS to 43 FPS. The additional trainable parameters mainly come from the wavelet transform encoder, which enhances long-range feature modeling with a lightweight overhead. The optimal transport module is used as a training-time matching loss and does not introduce additional inference parameters or change the memory-based inference pipeline. Therefore, SWT maintains practical inference efficiency while improving global matching capability and segmentation accuracy.
%===================================futher analysis===================================
\subsection{Further Analysis}
\label{subsec:further_ana}

\begin{figure*}[t]
      \centering
        \includegraphics[width=0.675\linewidth]{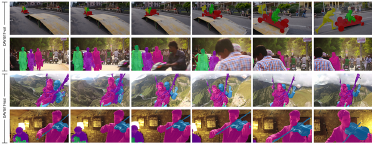}
        \caption{
            Qualitative results on the DAVIS dataset. From top to bottom, the first row, second row, third row and fourth row show qualitative results for DAVIS17-val and DAVIS17-test respectively. The temporal direction of video sequences proceeds from left to right.
        }
        \label{fig:davis}
    \end{figure*}

\begin{figure*}[t]
      \centering
        \includegraphics[width=0.675\linewidth]{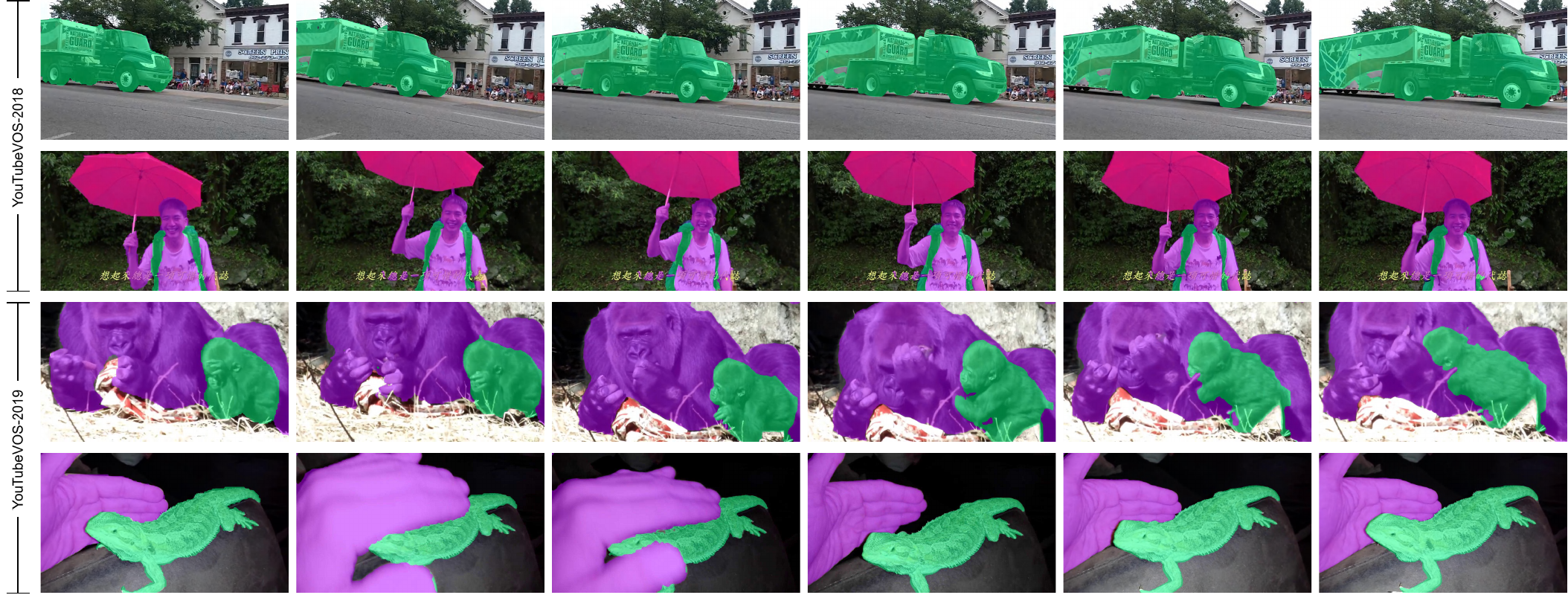}
        \caption{
            Qualitative results on the YouTubeVOS dataset. From top to bottom, the first and second rows show qualitative results for YouTubeVOS18-val, while the third and fourth rows display results for YouTubeVOS19-val. The temporal direction of video sequences proceeds from left to right.
        }
        \label{fig:ytb}
    \end{figure*}

\begin{figure}[t]
      \centering
        \includegraphics[width=0.441\linewidth]{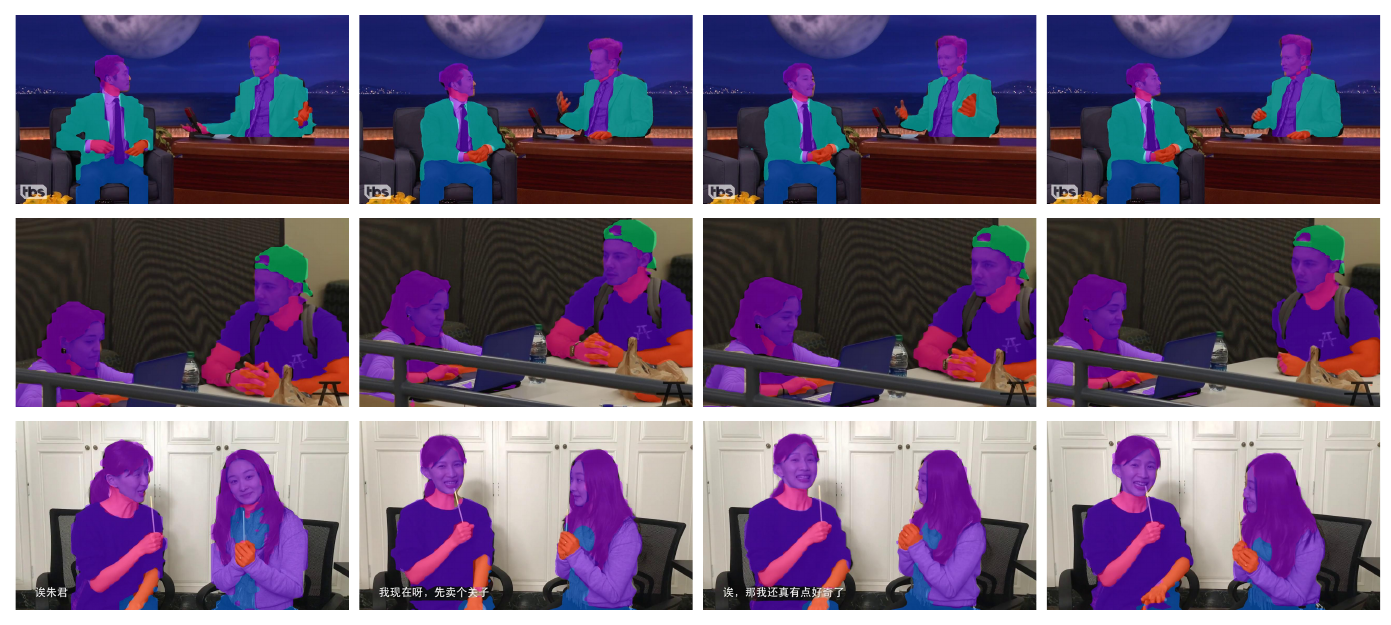}
        \caption{
            Qualitative results on the VIP dataset. The temporal direction of video sequences proceeds from left to right.
        }
        \label{fig:vip}
    \end{figure}

    Next, we conduct a more in-depth experimental analysis of each module, and we discuss how each module functions through a series of experimental results. Figures \ref{fig:davis}, \ref{fig:ytb}, and \ref{fig:vip} present qualitative results on DAVIS, YouTubeVOS, and VIP, respectively.

    \textbf{Sliding-Window Sample Buffer.} To analyze this scheduling strategy, Figure \ref{fig:ana_s} compares \(c=1,2,4,8\) while holding the buffer capacity at \(cN=64\); the corresponding refresh sizes are \(N=64,32,16,8\). All models are trained for 20 epochs under otherwise identical configurations. Table \ref{tab:ablation_sensitivity}(c) reports the final results of this fixed-capacity comparison together with the controlled \(N\) ablation at \(c=4\).

    \textbf{Wavelet Transform.} In order to demonstrate more intuitively the enhancement effect of the wavelet transform on the effective receptive field, we visually compare the effective receptive field of different architectures by calculating the gradient contribution of the input image to the model output centroid (Figure \ref{fig:ana_w} (c)). The experimental result shows that compared with the original ResNet-18 and ResNet-50, ResNet-18 improved with wavelet transform significantly enlarges the effective receptive field, which further validates the advantage of wavelet transform in feature extraction.

    \textbf{Optimal Transport.} In order to demonstrate the global matching mechanism based on optimal transmission more clearly, we design a set of visualization experiments. Specifically, we use a pre-trained backbone network to perform feature extraction on two randomly cropped images, subsequently compute the similarity matrix between the resulting features (e.g., Cross Correlation in Figure \ref{fig:main}), and the similarity matrix is visually represented by projecting it as a heatmap overlay onto the original input image. As shown in the yellow labeled area in Figure \ref{fig:ana_T_hm}, the experimental result represents that our method can effectively establish cross-image feature correspondences. To further analyze the contribution of local features to the global similarity, we also divide each image frame into a \(3 \times 3\) grid region (a total of 9 patches) and visualize the inter-patch correlation of the optimal transport capture through the similarity matrix (Figure \ref{fig:ana_T_cm}).

\section{Conclusion}
\label{sec:conclusion}
    In this paper, we propose a novel self-supervised VOS approach. Unlike previous approaches that require extensive pixel-level annotated video data for training, our method only needs to be trained on static images. A rolling sample buffer improves sample reuse across successive optimization steps. Additionally, we employ cascaded wavelet transforms to perform convolutions in the wavelet domain, significantly expanding the effective receptive field with moderate additional computational cost, thereby enhancing the long-range modeling capability of the convolutional architecture. Finally, we leverage optimal transport to achieve cross-frame global feature matching, addressing common challenges in VOS tasks such as rapid object motion and deformation. Our method achieves advanced results on multiple VOS benchmarks and even demonstrates competitive performance compared to some fully supervised VOS approaches.

\section{Acknowledgements}
This work is supported in part by the National Natural Science Foundation of China (Grant No. 62601561, 62176172, 61672364); partially by the National Key Research and Development Program of China (Grant No. 2018YFA0701701); partially by Basic Research Program of Jiangsu (Grant No. BK20250789).

%\section{Acknowledgments}

\bibliographystyle{ACM-Reference-Format}
\bibliography{sample-base}
%%
%% If your work has an appendix, this is the place to put it.
\appendix

\end{document}